\documentclass[conference]{IEEEtran}
\IEEEoverridecommandlockouts
\usepackage{colortbl}
\usepackage[compress]{cite}
\usepackage[hidelinks]{hyperref}
\usepackage{amsmath,amssymb,amsfonts}
\usepackage{algorithm}
\usepackage{algpseudocode}
\usepackage{graphicx}
\usepackage{textcomp}
\usepackage{xcolor}
\usepackage{subcaption}
\usepackage{amssymb}
\usepackage{amsmath}
\usepackage{cases}
\usepackage{graphics}
\usepackage{multirow}
\usepackage{multicol}
\usepackage{xspace}
\usepackage{array}
\usepackage{float}
\usepackage{fancyhdr}
\usepackage{orcidlink}

\def\BibTeX{{\rm B\kern-.05em{\sc i\kern-.025em b}\kern-.08em
    T\kern-.1667em\lower.7ex\hbox{E}\kern-.125emX}}

\makeatletter
\DeclareRobustCommand\onedot{\futurelet\@let@token\@onedot}
\def\@onedot{\ifx\@let@token.\else.\null\fi\xspace}

\def\eg{e.g\onedot} 
\def\ie{i.e\onedot}

\def\etal{et al\onedot}
\makeatother

\begin{document}
	\title{Hi-ALPS - An Experimental Robustness Quantification of Six LiDAR-based Object Detection Systems for Autonomous Driving\\
		\thanks{This work was accomplished within the project VAL, FKZ 5320000013, EBA Az. 8fd/003-1255\#008-VAL, funded by the German Federal Ministry of Digital and Transport.}
	}

\author{\IEEEauthorblockN{1\textsuperscript{st} Alexandra Arzberger}
\IEEEauthorblockA{\textit{Faculty of Computer Science} \\
\textit{Nuremberg Institute of Technology}\\
Nuremberg, Bavaria, Germany \\
alexandra.arzberger@th-nuernberg.de \\ 
0009-0001-2581-7119
}
\and
\IEEEauthorblockN{2\textsuperscript{nd} Ramin Tavakoli Kolagari}
\IEEEauthorblockA{\textit{Faculty of Computer Science} \\
\textit{Nuremberg Institute of Technology}\\
Nuremberg, Bavaria, Germany \\
ramin.tavakolikolagari@th-nuernberg.de \\
0000-0002-7470-3767
}
}



\maketitle

\begin{abstract}
Light Detection and Ranging (LiDAR) is an essential sensor technology for autonomous driving as it can capture high-resolution 3D data. As 3D object detection systems (OD) can interpret such point cloud data, they play a key role in the driving decisions of autonomous vehicles. Consequently, such 3D OD must be robust against all types of perturbations and must therefore be extensively tested. One approach is the use of adversarial examples, which are small, sometimes sophisticated perturbations in the input data that change, i.e., falsify, the prediction of the OD. These perturbations are carefully designed based on the weaknesses of the OD. The robustness of the OD cannot be quantified with adversarial examples in general, because if the OD is vulnerable to a given attack, it is unclear whether this is due to the robustness of the OD or whether the attack algorithm produces particularly strong adversarial examples.
The contribution of this work is \textit{ Hi-ALPS}---\underline{Hi}erarchical \underline{A}dversarial-example-based \underline{L}iDAR \underline{P}erturbation Level \underline{S}ystem, where higher robustness of the OD is required to withstand the perturbations as the perturbation levels increase. In doing so, the Hi-ALPS levels successively implement a heuristic followed by established adversarial example approaches. In a series of comprehensive experiments using Hi-ALPS, we quantify the robustness of six state-of-the-art 3D OD under different types of perturbations. The results of the experiments show that none of the OD is robust against all Hi-ALPS levels; an important factor for the ranking is that human observers can still correctly recognize the perturbed objects, as the respective perturbations are small. To increase the robustness of the OD, we discuss the applicability of state-of-the-art countermeasures. In addition, we derive further suggestions for countermeasures based on our experimental results.
\end{abstract}
	    

\thispagestyle{fancy}
\pagenumbering{gobble}
\renewcommand{\headrulewidth}{0pt}
\fancyhead{}
\fancyfoot[L]{\footnotesize This work has been accepted for publication in the IEEE Conference on
	Secure and Trustworthy Machine Learning (SaTML). The final version will
	be available on IEEE Xplore.}

\section{Introduction}
\label{sec:intro}
Autonomous driving offers several benefits, such as increasing the mobility of people who are unable to drive on their own, and improving the efficiency and safety of traffic. The recording of high-resolution sensor data of the environment, \ie, 3D data, is crucial for the safety of autonomous driving systems, since the perception sensors correspond to the eyes of the autonomous vehicle. Therefore, Light Detection and Ranging (LiDAR) is an essential sensor technology capable of capturing the environment of the autonomous vehicle as 3D point cloud data. 3D Object Detection systems (OD), mostly based on Machine Learning, \eg, \cite{pp, pointrcnn, second, voxset, pvrcnn, pvrcnnplus, ct3d, pdv}, are able to interpret such point cloud data by detecting objects contained in the point cloud scenes, \ie, localizing them and assigning a label. The 3D OD is of key importance for the driving decisions of the autonomous vehicle. Therefore, it must be carefully designed and extensively tested. Since LiDAR point cloud data is inherently noisy, inconsistently shaped, and prone to perturbations, \eg, occlusions, the OD must be robust against all kinds of perturbations, resulting from natural events (safety), \eg, different weather conditions such as rain, as well as perturbations generated by malicious attackers (security). Another approach besides adding common perturbations to the point cloud data, \eg, \cite{dong, robo3d, robuli}, is adversarial examples.
These correspond to small perturbations added to the input data that change the prediction of the OD \cite{ae}. There are various approaches to their generation, for point cloud classification, \eg, \cite{pcsm, xiang, liu, wick, geometryaware, lggan}, as well as for point cloud scenes, \eg, \cite{spoofsun, yang, scenesurvey, spoofcao}, by adding, dropping, shifting points, or even generative methods. However, these algorithms carefully craft adversarial examples, \ie, they optimize their perturbations based on observing the behavior of the OD towards the perturbations. On the one hand, this leads to effective attack algorithms; on the other hand, countermeasures lead to even more powerful attack algorithms \cite{3dcertify}. As a result, it is difficult to distinguish whether the robustness of an OD is insufficient or whether the attack algorithm has created a strong adversarial example.

 In order to quantify the robustness of state-of-the-art OD, we propose \textit{Hi-ALPS} (\underline{Hi}erarchical \underline{A}dversarial-example-based \underline{L}iDAR \underline{P}erturbation Level \underline{S}ystem), a perturbation level system, where with increasing level, a higher robustness of the OD is required to withstand the perturbations (Section \ref{sec:pertlevel}). Hi-ALPS  approaches the heuristic that outer points  have the highest impact on the prediction of the OD, because these define the object's contour \cite{pcsm}.  Since our perturbations are based on a heuristic instead of model internal information, \eg, the gradient, our experiments correspond to a black-box scenario. We evaluate six state-of-the-art OD with Hi-ALPS, which cover various categories, based on their used data representation as well as their architecture. To evaluate the robustness against the perturbations included in Hi-ALPS, we present metrics for robustness evaluation as well as for measuring the perceptibility of the perturbations, since adversarial examples shall be unobtrusive \cite{ae}. 
  Our results show none of the investigated OD to be robust against the Hi-ALPS levels, except for the first Hi-ALPS level that includes shifting random points into random directions. Thereby, the investigated OD do not even show a sufficient performance on natural data, corresponding to Hi-ALPS level 0. With increasing Hi-ALPS level, the robustness of all OD decreases. 
 In summary, our key contributions are:
\begin{itemize}[\IEEEsetlabelwidth{Z}]
	\item Hi-ALPS, a hierarchical perturbation level system, where with increasing level the perturbations approach a heuristic and a higher robustness of the OD is required to withstand the perturbations (Section \ref{sec:pertlevel}). 
	All investigated perturbations correspond to a black-box scenario. Our experiments show  the OD to be not robust against our generated perturbations, except for Hi-ALPS level 1 (Section \ref{sec:res}). Even for Hi-ALPS level 0, including the evaluation on natural data without any perturbations, the OD do not show a sufficient performance.
	\item AELiDAR, a prototypical validation framework, in order to generate perturbed point cloud data, according to our defined Hi-ALPS levels (Section \ref{sec:pertimpl}).
	\item With Hi-ALPS, we quantify the robustness of six state-of-the-art OD. Our experimental validation shows the OD to be truly not sufficiently robust, because the investigated perturbations do not correspond to strong adversarial examples against which the OD has no chance, but a heuristic instead.
\end{itemize}
Since the OD are not robust already against Hi-ALPS perturbation level 2, first the training process should be optimized instead of applying complex countermeasures. 
Nevertheless, we discuss the applicability of existing countermeasures from related work and propose further suggestions for countermeasures based on our experimental validation in order to increase the robustness of OD against higher Hi-ALPS levels.

\section{Related Work}
\label{sec:relatedwork}
The following provides related work, including basics about 3D object detection and point cloud perturbations. 

\subsection{3D Object Detection} \label{sec:3dod}
\begin{figure}[!t]
	\centering
	\includegraphics[width=0.45\textwidth]{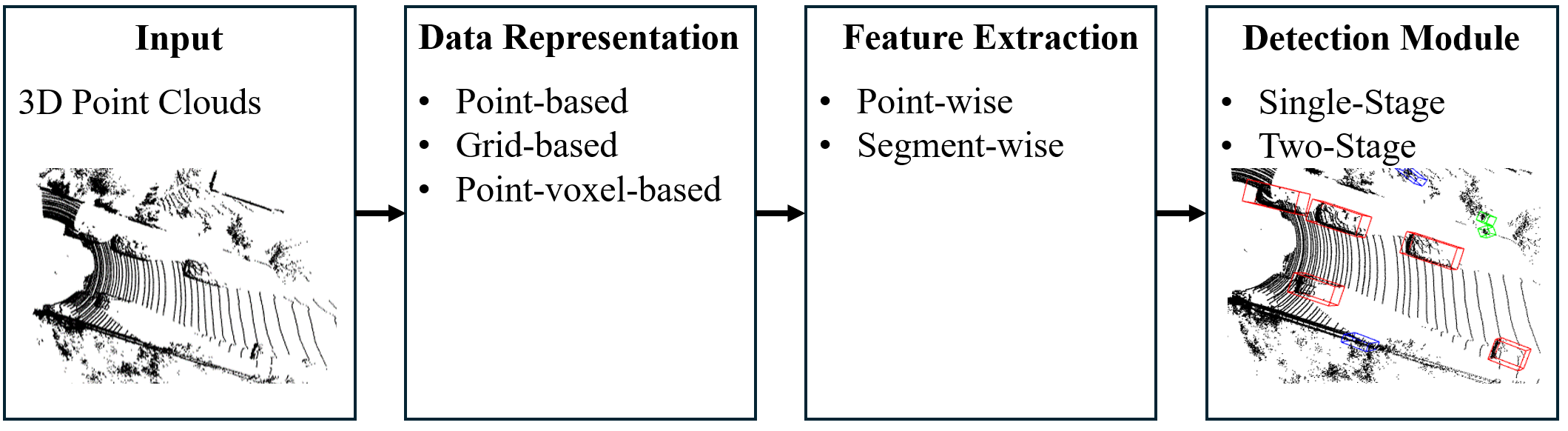}
	\caption{Pipeline of state-of-the-art LiDAR-based 3D object detection systems \cite{drob}.}
	\label{fig:func3d}
\end{figure}

3D OD are capable of interpreting point cloud scenes through the localization and classification of point cloud objects. Figure \ref{fig:func3d} shows the systematization of the  pipeline of 3D OD according to Drobnitzky \etal \cite{drob}, regarding different design choices addressing the used representation and architecture of the OD. Thereby, the input data corresponds to a 3D point cloud scene. Since 3D point cloud data is sparse and irregular, a suitable representation is required to properly process the input. Based on the used representation, recent publications, \eg, \cite{mao, scenesurvey}, divide 3D OD into three categories, \ie, \textit{point-based},  \textit{grid-based}, and \textit{point-voxel-based}. Thereby, point-based methods, \eg, \cite{pointnet, pointrcnn}, directly consume the raw point cloud data, while grid-based, \eg, \cite{parta, pp, second, voxset}, divide the point cloud into a grid structure. 
Point-voxel-based strategies, \eg, \cite{pvrcnn}, combine point-based and grid-based representation, whether through different detection stages or the fusion of point and voxel features. After the computation of the representation, the feature extraction follows.  Thereby, the dimension of the data is reduced to compute a robust set of features. According to Drobnitzky \etal, there exist two categories for the feature extraction for LiDAR-based OD, point-wise and segment-wise. Point-wise approaches are realized, \eg, by PointNet \cite{pointnet} and PointNet++ \cite{pointnetplus}, while segment-wise are based, \eg, on 2D or 3D convolutional neural networks (CNN).
 The final stage of the pipeline is the detection module, which localizes and classifies the point cloud objects, \ie, computing bounding boxes (BB) and assigning a label. There exist two types for the architecture of the detection module, single-stage and two-stage architectures. Thereby, two-stage architectures, \eg, \cite{parta, pointrcnn, pvrcnn, voxset}, first propose regions that are likely to contain an object, which are refined in a second step. In contrast, single-stage architectures, \eg, \cite{pp, second, voxset}, directly perform the prediction. Table \ref{tab:odcat} contains an overview of various state-of-the-art 3D OD and their categorization regarding the used representation and architecture of the detection module.

\begin{table}[t!]
	\centering
	\caption{Categorization of state-of-the-art 3D object detection systems based on their used representation and architecture \cite{scenesurvey}.}
	\begin{tabular}{p{1.77cm}| >{\centering}p{0.3cm} >{\centering}p{0.3cm} >{\centering}p{1.35cm}| >{\centering}p{1.45cm} >{\centering\arraybackslash}p{1.25cm}}
		\hline
		Detector & \multicolumn{3}{c|}{Representation} & \multicolumn{2}{c}{Architecture} \\
		& Point & Grid & Point-Voxel &
		Single-Stage       & Two-Stage       \\ \hline 
		Part-A² \cite{parta}&       & \checkmark      &             &                        &            \checkmark      \\ 
		PointPillars \cite{pp} &       &   \checkmark    &             &                    \checkmark     & \\
		PointRCNN \cite{pointrcnn}& \checkmark                   &       &                  & & \checkmark \\
		PV-RCNN \cite{pvrcnn} &       &       &   \checkmark          &       &                \checkmark \\
		SECOND \cite{second}&       &   \checkmark    &             &              \checkmark          & \\
		VoxSeT \cite{voxset}&       &    \checkmark   &             &                \checkmark       & \checkmark \\	\hline
	\end{tabular}
	\label{tab:odcat}
\end{table}

\subsection{Point Cloud Perturbations} \label{sec:pert}
In this section, we present related work about point cloud perturbations. These include common corruptions and adversarial examples. Finally, related work about countermeasures to increase the robustness of the 3D OD against these follows.

\subsubsection{Common Corruptions} \label{sec:corrup}
Common corruptions correspond to rather random perturbations that may occur in real-world data due to natural circumstances, \eg, different weather conditions, such as fog or snow \cite{robuli, dong, robo3d}, Gaussian noise \cite{robuli, dong} or motion blur \cite{dong, robo3d}. 
Unlike adversarial examples, these perturbations are not specifically designed to alter the prediction of the OD, \ie, they are not optimized to affect the performance of the OD. Instead, they aim to simulate natural scenarios to investigate the robustness of the OD.

\subsubsection{Adversarial Examples} \label{sec:ae}
Adversarial examples correspond to small perturbations added to the input data intended to change the prediction of the OD \cite{ae}. 
The following presents related work for the generation of adversarial examples for point cloud classification, \ie, perturbing single point cloud objects,  and for LiDAR point cloud scenes for object detection in context of autonomous driving. 

\paragraph{Point Cloud Objects}
According to Naderi and Baji\'{c} \cite{classisurvey}, adversarial point cloud objects can be generated by \textit{adding}, \eg, \cite{minimal, xiang}, \textit{dropping}, \eg, \cite{wick, pcsm}, \textit{shifting}, \eg, \cite{minimal, ext, advshape, geometryaware, xiang}, points or \textit{transforming}, \eg, through generative approaches, such as \cite{advpc, lggan}.   
Thereby, Xiang \etal \cite{xiang} are the first to generate adversarial examples on 3D point cloud data by adding and shifting points using an optimization problem based on the gradient. Consequently, their algorithm corresponds to a white-box attack. 
In contrast, Zheng \etal \cite{pcsm} generate adversarial examples by dropping points from the point cloud object. Their algorithm PointCloud Saliency Maps (PCSM) shifts outer points to the point cloud center, assuming that these are the most influential points for the prediction of the 3D OD. Thereby, PCSM calculates a saliency score that represents the influence of each point on the gradient and  generates the adversarial example by dropping the most influential points, \ie, the points with the highest saliency score. 
Generative approaches are realized through neural networks, which are capable of generating data. Thereby, Hamdi \etal \cite{advpc} use an autoencoder for generating adversarial point cloud objects, while Zhou \etal \cite{lggan} employ a Generative Adversarial Network (GAN). 

\paragraph{LiDAR Point Cloud Scenes}
A popular approach for generating adversarial point cloud scenes is LiDAR spoofing, \ie, generating adversarial points with additional laser beams, whether added in the real world or algorithmic. Cao \etal \cite{spoofcao} apply LiDAR spoofing based on an optimization algorithm to add fake obstacles in front of the victim vehicle to make it disappear for the OD.
The approach of Sun \etal \cite{spoofsun} is based on the observation that far-away obstacles and occluded vehicles contain smaller numbers of points. They extract occluded as well as far-away vehicles from the validation split of KITTI dataset and further create their own adversarial point objects to spoof them into the natural point cloud scene. A robust OD should not identify these fake objects as real, since these violate physical occlusion laws. 
Tu \etal \cite{physae, tu22} place adversarial meshes as roof luggage on vehicles under realistic conditions, \ie, considering occlusion and environmental lightning, in order to fool multi-sensor perception systems including camera and LiDAR.
Yang \etal \cite{yang} randomly add adversarial points above the victim vehicle and optimize their position based on the gradient with the goal to make the victim object invisible for the 3D OD.
Sato \etal \cite{sato} investigate the capabilities of spoofing attacks in the context of different datasets, LiDAR sensors, and OD. With their spoofing attack, they are able to spoof  6,000 points. 
An example of shifting points is the publication of Wang \etal  \cite{advperturbscene}, who perturb point objects of interest, \ie, cars pedestrians, and cyclists. Therefore, they use a dual loss function to generate adversarial examples, which consists of an adversarial loss, for optimizing the perturbation to change the prediction, and a perturbation loss, for limiting the perturbations.
Zhang \etal \cite{scenesurvey} generate adversarial LiDAR point cloud scenes from the KITTI \cite{kitti} and Waymo \cite{waymo} datasets through shifting, dropping, and adding points. 
Thereby, they formulate the shift attack as a gradient-based optimization problem. Their method for adding points is based on the publication of Xiang \etal \cite{xiang} and the method for dropping points is based on PCSM by Zheng \etal \cite{pcsm}. Thereby, Zhang \etal found the 3D OD to be the most vulnerable against shifting points, while the perceptibility of this attack is the lowest.

The disadvantage of white-box attack algorithms, \eg, \cite{xiang, pcsm, scenesurvey}, lies in their dependence on the gradient, where access cannot be guaranteed in the real world. In contrast, generative methods, \eg, \cite{lggan, advpc}, require training with large amounts of train data and their transferability to real-world scenarios is limited. Spoofing attacks, \eg, \cite{spoofcao, spoofsun}, are generally physically realizable, but in practice, it is probably rather difficult to implement the additional laser beams to spoof the points in the positions simulated with the code. 

In summary, all these publications investigate the robustness of 3D OD against sophisticated perturbations, which aim to change the prediction. Consequently, it is unclear, whether the OD are truly not robust or if the implemented perturbations correspond to particularly strong attacks.
In contrast, the perturbations included in Hi-ALPS approach a heuristic, where with increasing level a higher robustness of the OD is required to withstand them. Therefore, Hi-ALPS enables the quantification of the robustness of the OD. 

\subsubsection{Countermeasures}
Naderi and Bajić \cite{classisurvey} divide countermeasures against
 perturbations on point cloud data into data-focused and model-focused methods. The data-focused methods refer to countermeasures that address the input data of the OD.
 These can be further divided into input transformations, which correspond to applying transformations to the input point cloud data as a preprocessing step in order to desensitize the OD against the perturbations or training data optimization, \ie, adapting the train data to optimize the knowledge of the OD.
 Model-focused strategies are further divided into deep model modification, model retraining, and combined strategies. Thereby, deep model modification corresponds to adapting the architecture of the OD and may include retraining it.  Model retraining is similar to training data optimization, but  involves further strategies than including adversarial data to the training data.  Combined strategies combine both, deep model modification and retraining. 
The following includes state-of-the-art countermeasures to improve the robustness of the OD against perturbations, regarding point cloud classification as well as methods for LiDAR point cloud scenes.
 
 \paragraph{Point Cloud Objects}
 Regarding training data optimization, one approach lies in data augmentation corresponding to training the model with an extended dataset, \eg,  \cite{robuli, dong}. 
Another approach is adversarial training, proposed by Goodfellow \etal \cite{advtrain}, where adversarial examples assigned with the correct labels are added during training, \eg, \cite{tu22, scenesurvey}.
Furthermore, there exist various approaches for input transformations. Thereby, simple random sampling randomly drops a fixed part of points from the point cloud with equal probability, in order to catch and remove adversarial points \cite{classisurvey}. Statistical outlier removal removes points, whose distance to other points exceeds the average distance \cite{robuli, liu, scenesurvey}. DUP-Net (Denoiser and Upsampler Network), proposed by Zhou \etal \cite{dupnet}, first applies statistical outlier removal to denoise the data and afterward aims to restore the point cloud by upsampling the remaining points. Another approach is salient point removal \cite{ext}, which is based on the assumption that adversarial points are in particular influential for the gradient, since these change the prediction. Therefore, salient point removal first determines the most influential points based on observing the gradient and afterward removes them. 

\paragraph{LiDAR Point Cloud Scenes}
For LiDAR scenes, there exist countermeasures exploiting physical properties, such as point distances and occlusions. Sun \etal \cite{spoofsun} present CARLO---o\underline{C}clusion-\underline{A}ware hie\underline{R}archy anoma\underline{L}y detecti\underline{O}n, against their spoofing attack (Paragraph \ref{sec:ae}), which identifies spoofed fake points based on physical laws related to occlusion. They also propose Sequential View Fusion (SVF), which exploits the fact that occlusions are particularly noticeable from the front view. It includes a semantic segmentation network processing the front view of the LiDAR scene and a view fusion module to apply the prediction of the semantic segmentation network to the input data of the 3D OD. 
The countermeasure proposed by Yang \etal 
\cite{yang} refers to the distance between adjacent laser beams, because this differs for adversarial points compared to natural points.
\section{AELiDAR: Exploring the Robustness of 3D Object Detection Systems based on Adversarial Examples}
The experimental setup \textit{AELiDAR} (Adversarial Examples for Light Detection and Ranging) serves to evaluate the robustness of 3D OD based on the approach of adversarial examples. It includes various perturbation methods to realize different perturbations included in the hierarchical perturbation level system Hi-ALPS (Section \ref{sec:pertlevel}). The following includes the perturbation level system to be realized with AELiDAR, the robustness evaluation criteria, and the implementation details of the realized perturbations.

\subsection{Hi-ALPS---A Hierarchical Perturbation Level System for Robustness Quantification} \label{sec:pertlevel}
We propose Hi-ALPS (\underline{Hi}erarchical \underline{A}dversarial-example-based \underline{L}iDAR \underline{P}erturbation Level \underline{S}ystem), a level system for the quantification of the robustness of LiDAR-based OD against perturbations, \ie, the identification of the threshold at which the OD is not robust anymore. 
With an increasing perturbation level, we assume a greater robustness of the OD is required to withstand the perturbation, \ie, the performance of a not robust OD decreases with increasing level, as confirmed in Section \ref{sec:pertres}. Hi-ALPS level 0 includes the performance on natural data, \ie, there are no perturbations applied to the data. Hi-ALPS level 1 includes random perturbations. Between Hi-ALPS levels 2 to 5, the perturbations approach a heuristic,
 while Hi-ALPS level 6 corresponds to an arbitrary adversarial example algorithm, \eg, \cite{scenesurvey, spoofsun, sato}, considering OD-specific weaknesses. 
Note that Hi-ALPS levels 1 to 5 correspond to a black-box approach and do not require training or an optimization process, whereas this does not necessarily apply for Hi-ALPS level 6.  

\begin{figure}[!t]
	\centering
	\includegraphics[width=0.5\textwidth]{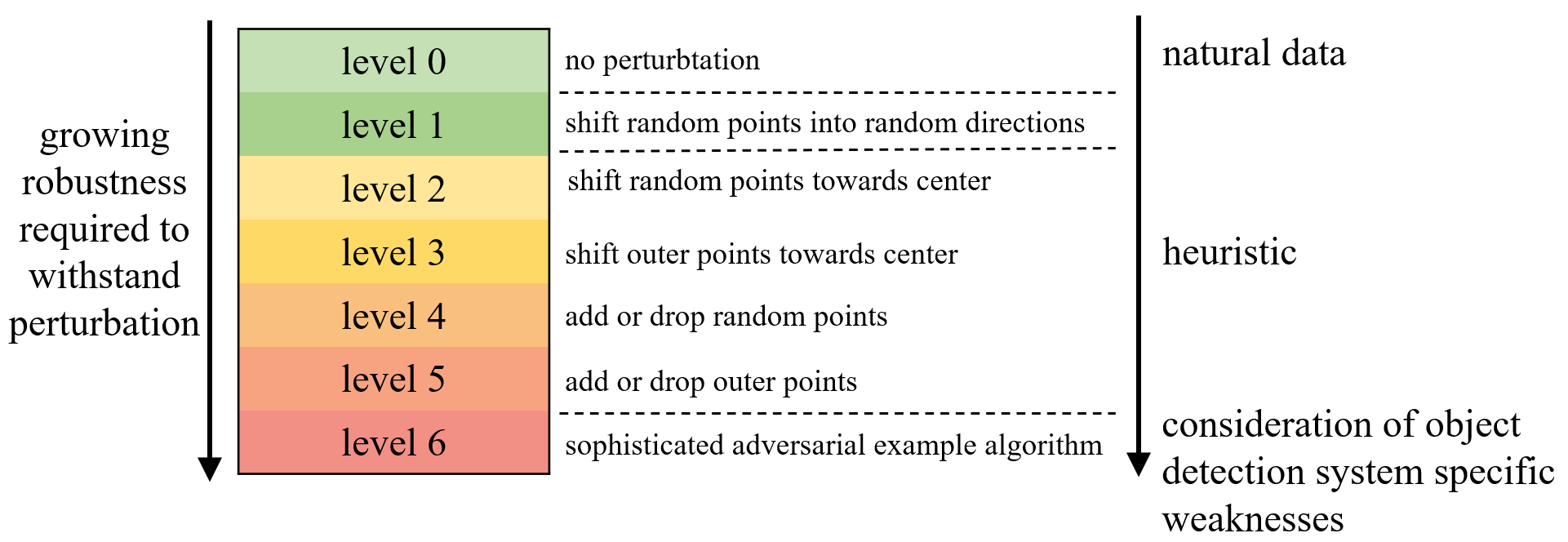}
	\caption{Systematization of Hi-ALPS.}
	\label{fig:pert}
\end{figure}

 We realize Hi-ALPS levels 1 to 5, excluding Hi-ALPS level 6, because of the large variety of approaches for its realization. 
Thereby, we approach the only heuristic considered applicable based on our current understanding of the research literature that outer points are in particular influential, since they define the contour of the objects \cite{pcsm, occam}. Figure \ref{fig:pert} includes the perturbations contained in the Hi-ALPS levels. 
 In Hi-ALPS level 1, we shift random points into random directions, since this corresponds to a random perturbation, where the number of included points remains the same. We define Hi-ALPS level 2 as shifting random points into a determined direction, \ie, towards the center or away. The shift towards a determined direction may already lead to a perturbation of the contour. We shift random points towards the center because this directly approaches the heuristic. Furthermore, shifting points away from the center will probably increase the perceptibility of the perturbations, since the contour is more likely to be perturbed. This change of the contour is likely to influence the OD and, therefore, the perturbation may be too harmful to be considered as only Hi-ALPS level 2.  Hi-ALPS level 3 corresponds to shifting points towards the opposite direction. Based on the reasons behind our selection for Hi-ALPS level 2, we shift outer points towards the center. In Hi-ALPS levels 4 and 5, we perturb the point cloud data by changing the number of included points, by adding, comparable to spoofing attacks, \eg, \cite{spoofcao, spoofsun}, as well as by dropping points, \eg, \cite{pcsm,scenesurvey}. In Hi-ALPS level 4, the perturbation is random. In contrast, Hi-ALPS level 5 corresponds to the perturbation of the point number in determined locations. Here, we perturb the point cloud contour by adding or dropping outer points to fully approach the heuristic. 
 
\subsection{Evaluation Criteria}
\label{sec:robustnessmetr}
In the following, we present our metrics for the robustness evaluation of the OD and the evaluation of Hi-ALPS. These include on one hand robustness evaluation metrics, which refer to the robustness against the investigated perturbations. On the other hand, adversarial examples are defined as imperceptible perturbations (Paragraph \ref{sec:ae}). Therefore, we measure the perceptibility of the perturbations as well. 

\subsubsection{Robustness Evaluation Metrics}
For robustness evaluation, we use the Average Precision Ratio $\mathit{AP \: ratio}$, the False Negative Attack Success Rate $\mathit{FN\_ASR}$ and the False Positive Attack Success Rate $\mathit{FP\_ASR}$.

The AP ratio corresponds to the ratio of the AP values of one class on natural and perturbed data. The mAP ratio corresponds to the ratio of the mean value of all class  AP values \cite{scenesurvey}. According to the KITTI evaluation \cite{kittibench}, the thresholds for the Intersection over Union (IoU) are $\mathit{IoU}\geq 0.5$, for the classes pedestrian and cyclist, and $\mathit{IoU} \geq 0.7$, for the class car. The AP ratio can take values between 0 to 1, where $AP=1$ signifies that the performance did not change on the perturbed data, \ie, the OD is perfectly robust. 

 We adopt the commonly used metric Attack Success Rate (ASR) in the sense of Wang \etal \cite{advperturbscene} to measure the fraction of objects per scene that were detected before the perturbation, but were not detected afterward. Following Tu \etal \cite{tu22}, we denote this metric as False Negative Attack Success Rate \textit{FN\_ASR}.  We use the IoU thresholds defined by the KITTI evaluation \cite{kittibench}.
Equation \ref{eq:fn} shows the calculation of the FN\_ASR for the class car referring to a natural point cloud scene $P$, containing a set of cars $\mathit{Car}^P$, where $|\mathit{Car}|^P$ is the number of cars, and its corresponding perturbed scene $P'$. For the calculation, a correction factor $C$ is summed up. This factor is $C=0$ if the IoU threshold is achieved between the BB of the $n$-th car $\mathit{BB}_{\mathit{Car}_n}$ in the natural point cloud scene $P$ and $\mathit{BB}_{\mathit{Car'}_n}$, the BB of the corresponding car detected in the perturbed point cloud scene $P'$,
or $C=1$ otherwise.
Note that the FN\_ASR is only defined for scenes, where at least one object was detected on natural data, because otherwise no objects can disappear.
\begin{equation}\label{eq:fn}
	\mathit{FN\_ASR@70}_{\mathit{Car}} (P, P')= \frac{\sum\limits_{n=0}^{|Car|^{P}} C@70_{\mathit{Car}_n}}{|\mathit{Car}|^{P}} 
\end{equation}

\begin{subnumcases}
	{C@70_{\mathit{Car}_n} =} 
	0, & if $\mathit{IoU}(\mathit{BB}_{\mathit{Car}_n}, \mathit{BB}_{\mathit{Car'}_n})  \geq 0.7 $\\
	1, & else
\end{subnumcases}

The FP\_ASR corresponds to the fraction of objects detected in a perturbed scene $P'$, which were not detected on natural data. Tu \etal \cite{tu22} use FP\_ASR as well, but they refer to scenes in which at least one false positive prediction appeared.
Equation \ref{eq:fpcar} corresponds to the FP\_ASR for the set $\mathit{Car'}^{P'}$, which includes all cars detected in the adversarial point cloud scene $P'$, where $|\mathit{Car'}|^{P'}$ is the number of cars included in the set. As well as for FN\_ASR, a correction factor $C$ is summed up. Thereby, $C=0$, if the $n$-th car was already detected on natural data and $C=1$, otherwise. 
\begin{equation}\label{eq:fpcar}
	\mathit{FP\_ASR@70}_{\mathit{Car}} (P, P')= \frac{\sum\limits_{n=0}^{|Car'|^{P'}} C@70_{\mathit{Car'}_n}}{|\mathit{Car'}|^{P'}} \\
\end{equation}

\begin{subnumcases}
	{C@70_{\mathit{Car'}_n} = }
	0, & if $\mathit{IoU}(\mathit{BB}_{\mathit{Car'}_n}, \mathit{BB}_{\mathit{Car}_n}) \geq 0.7$ \\
	1, & else
\end{subnumcases}
Both metrics, FN\_ASR and FP\_ASR, can take values from 0 to 1. The higher the value of the FN\_ASR, the more objects were overlooked by the OD on perturbed data. The higher the FP\_ASR, the more false positive detections happened, \ie, objects detected on perturbed data that were not detected on natural data. Consequently, the higher the values of FN\_ASR and FP\_ASR the lower is the robustness of the OD against the perturbations.
In addition to the classes car, pedestrian, and cyclist, we calculate the FN\_ASR and FP\_ASR for all objects, without consideration of the predicted class, \eg, a pedestrian counts as detected even if it was falsely classified as a car. For this, we chose an IoU threshold of $\mathit{IoU}\geq 0.5$.
To evaluate the robustness across all perturbed point cloud scenes,  we compute the mean values of FN\_ASR and FP\_ASR. 

\subsubsection{Perturbation Metrics}
For measuring the perceptibility of the perturbations, we use the Perturbation Rate $\mathit{PR}$, the Chamfer and Hausdorff distance.
The PR corresponds to the fraction of the number of points to be perturbed. 
Both, Chamfer and Hausdorff distance are based on the Euclidean distance. Equation \ref{eq:eucl} corresponds to the calculation of the Euclidean distance between the i-th point $p^O_i$ inside a natural point cloud object $P_O$ and the j-th point $p'^O_j$ in an adversarial object $P'_O$. Note that even though if LiDAR points contain a fourth coordinate, \ie, the intensity of the reflected laser beam, we only consider three dimensions in the calculation of our distance metrics, since we do not perturb the intensity value, consequently, $\mathit{p}_{i}, p'_j  \in \mathbb{R}^{3}$. 
\begin{equation}\label{eq:eucl}
	\begin{split}
		D_{E} (p^O_i, p'^O_j) = ||p^O_i - p'^O_j||_2 = \\
		\sqrt{((p^O_{i,x} - p'^O_{j, x})^2 + (p^O_{i, y} - p'^O_{j, y})^2 + (p^O_{i, z} - p'^O_{j, z}))^2}
	\end{split}
\end{equation}

Equation \ref{eq:cham} adopts the definition of Kim \etal \cite{minimal} for Chamfer distance $D_C$. It corresponds to the mean Euclidean distance (Equation \ref{eq:eucl}) of every natural point $p$ included in a natural point cloud object $P_O$ to the nearest included point $p'$ in the corresponding perturbed point cloud object $P'_O$ and the other way around to consider potential far away points included in the perturbed data. The Chamfer distance corresponds to the maximum value of both calculations. 
\begin{equation}\label{eq:cham}
	\begin{split}
		D_C(P_O, P'_O) = \max \{ \frac{1}{|P_O|} \sum\limits_{p^O_i \in P_O} \min\limits_{p'^O_j \in P'_O} || p^O_i - p'^O_j||_2, \\ \frac{1}{|P'_O|} \sum\limits_{p'^O_j \in P'_O} \min\limits_{p^O_i \in P_O} ||p'^O_j - p^O_i||_2 \}
	\end{split}
\end{equation}

Equation \ref{eq:haus} adopts the calculation of Hausdorff distance from  Kim \etal \cite{minimal}.
In contrast to Chamfer distance, the Hausdorff distance corresponds to the maximum value of the Euclidean distances between a natural point $p_i$ in $P_O$ and the nearest point $p'_j$ in the perturbed point cloud object $P'_O$.
\begin{equation}\label{eq:haus}
	\begin{split}
		D_H(P_O, P'_O) = \max\{ \max\limits_{p^O_i \in P_O} \{ \min\limits_{p'^O_j \in P'_O} || p^O_i-p'^O_j ||_2 \} , \\ \max\limits_{p'^O_j \in P'_O} \{ \min\limits_{p^O_i \in P_O}\} || p'^O_j -p^O_i ||_2\} \} 
	\end{split}
\end{equation}
The higher the Chamfer and Hausdorff distance, the higher is the perceptibility of the perturbations. To measure the perceptibility across all  point cloud scenes, we compute the mean Chamfer and Hausdorff distance for all perturbed objects.

\subsection{Implementation of AELiDAR} \label{sec:pertimpl}
With our experimental setup AELiDAR, we perturb the validation split of KITTI \cite{kitti} dataset $\mathcal{P}_{\mathit{KITTI}}$. We investigate the robustness of six pretrained state-of-the-art 3D OD including Part-A² \cite{parta}, PointRCNN \cite{pointrcnn}, PointPillars \cite{pp}, PV-RCNN \cite{pvrcnn}, SECOND \cite{second} and VoxSeT \cite{voxset}, provided by the open source toolbox OpenPCDet \cite{opd} on perturbed KITTI validation split, \ie,  Hi-ALPS level 0 is already conducted. 
 Algorithm \ref{alg:levels} serves to generate the perturbed dataset $\mathcal{P'}_{\mathit{KITTI}}$.
 Since the perturbations include a random component, we conduct our tests in three iterations.
We apply the Hi-ALPS levels to single point cloud objects of interest $P_O$, \ie, the objects, which the OD is capable to detect.
Since these objects represent humans,  they are the most critical objects for the driving decisions. AELiDAR extracts the point objects of interest detected by the OD based on the predicted BB and perturbs all of them, where $\mathit{PR} \cdot |P_O| \geq 1$, \ie, AELiDAR can perturb at least one point. 
We conduct  an additional experiment in which AELiDAR expands the BB of each point object of interest on each side by a fraction $\mathit{env}=0.3$ of the length of the room diagonal of the BB, in order to investigate the influence of the object's environment. Thereby, the bottom of the BB is not extended, since it is limited by the ground. In the vertical direction, we half the extension, 
because the horizontal extend of the point cloud data is larger than the vertical.

\begin{figure}[t!]
	\centering
	\begin{subfigure}{0.1\textwidth}
		\includegraphics[width=\linewidth]{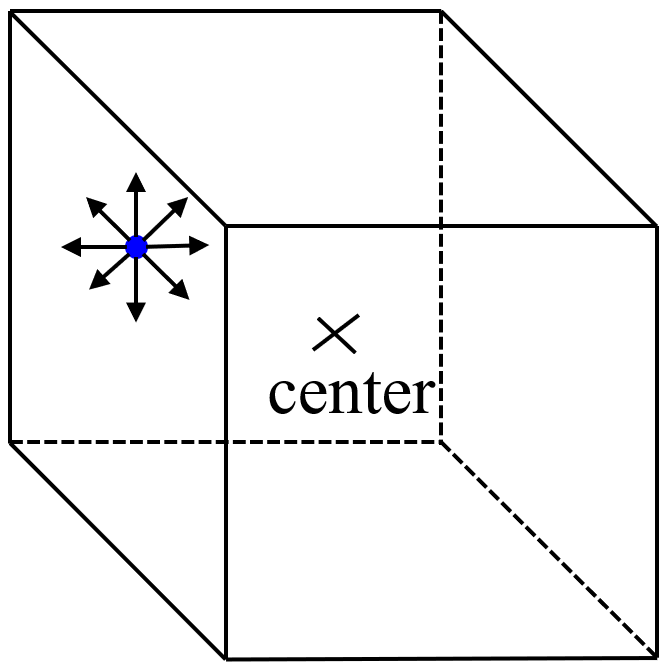}
		\caption{shift random} \label{fig:1a}
	\end{subfigure}
	\begin{subfigure}{0.1\textwidth}
		\includegraphics[width=\linewidth]{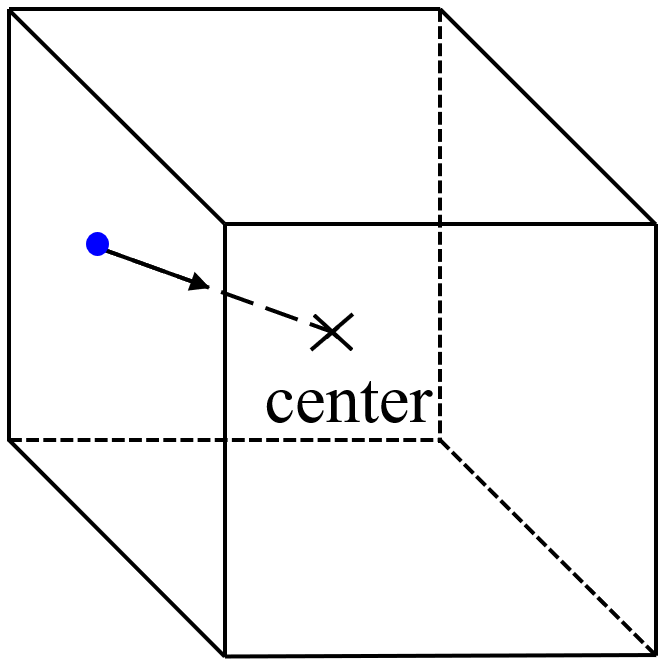}
		\caption{shift towards center} \label{fig:1b}
	\end{subfigure}
	\begin{subfigure}{0.1\textwidth}
		\includegraphics[width=\linewidth]{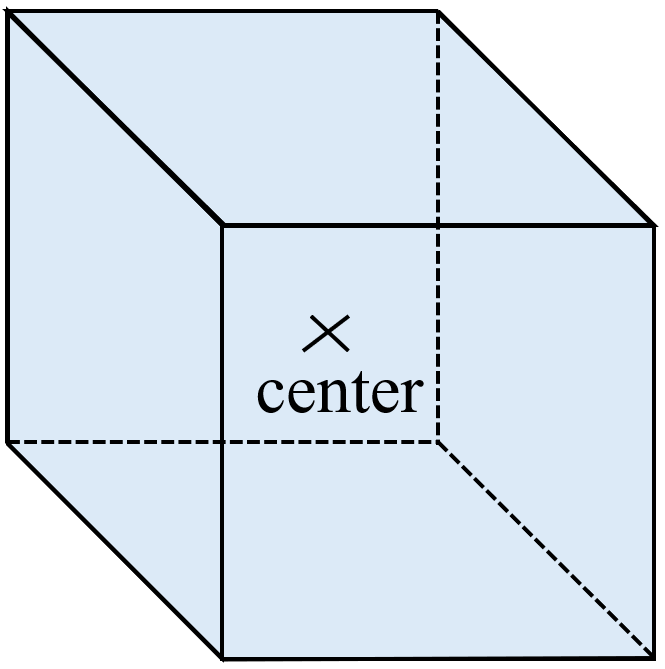}
		\caption{add random} \label{fig:1c}
	\end{subfigure}
	\begin{subfigure}{0.1\textwidth}
		\includegraphics[width=\linewidth]{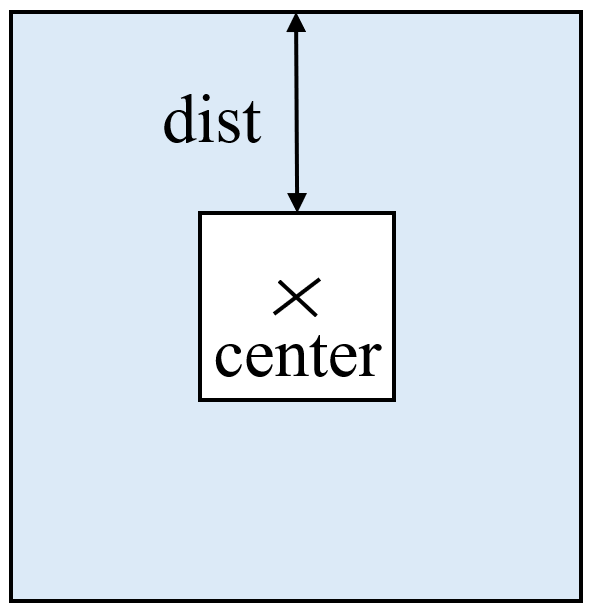}
		\caption{add outer} \label{fig:1d}
	\end{subfigure} 
	
	\caption{Perturbations generated by AELiDAR to realize the Hi-ALPS levels. }
	\label{fig:pertfunc}
\end{figure}

If no novel points are added, AELiDAR chooses points to be shifted or dropped, 
depending on the Hi-ALPS level. For Hi-ALPS levels 1, 2, and 4, AELiDAR selects random points be to perturbed. For Hi-ALPS levels 3 and 5, it chooses outer points. These correspond to the points with the largest Euclidean distance towards the center of the predicted BB. 
Figure \ref{fig:pertfunc} illustrates the perturbations generated by AELiDAR. For Hi-ALPS level 1, AELiDAR shifts points into random directions, as illustrated in Figure \ref{fig:1a}, by replacing the coordinates of the selected points with random coordinates. The maximum shift distance $\mathit{MSD}$ between the original point and the perturbed point is limited by the BB and the shift factor (SF), \ie, $\mathit{MSD} = \mathit{SF} \cdot d$, where $d$ is the length of the room diagonal.
Hi-ALPS levels 2 and 3 shift points towards the BB center. For traceability, AELiDAR shifts the selected points along the straight connecting the point and the center of the BB (Figure \ref{fig:1b}). The shift distance is only limited by the center and not additionally by the SF. 

The locations where AELiDAR adds points, are marked blue in Figure \ref{fig:pertfunc}. The intensity of added points corresponds to the value of the nearest natural point. When adding random points, AELiDAR generates them anywhere inside the BB (Figure \ref{fig:1c}). For outer points, AELiDAR adds points inside the outer area of the BB (Figure \ref{fig:1d}). Thereby, the width $\mathit{dist}$ of this area is limited by the SF and depends on the length $l$, width $w$, and height $h$ of the BB, $\mathit{dist} = \mathit{SF} \cdot \min{(l, w, h)} $. AELiDAR uses the minimum of $(l, w, h)$ to prevent opposing areas from overlapping. 
\begin{algorithm}[h!]
	\caption{Generate perturbed dataset $\mathcal{P}'_{\mathit{KITTI}}$}
	\label{alg:levels}
	\begin{algorithmic}[1]
		\Require $\mathcal{P}_{\mathit{KITTI}}$, predicted BB, configurations (\eg, Hi-ALPS level, $PR$)
		\State $\mathit{iteration} = 0$
		\For{$\mathit{iteration} < 3$}
		\For{$P$ in $\mathcal{P}_{\mathit{KITTI}}$}
		\For{$P_O$ in $P$}
		\If{$\mathit{env} > 0$}
		\State Expand BB
		\EndIf
		\State Extract $P_O$ from $P$ based on BB coordinates
		\If{$\mathit{PR} \cdot |P_O| \geq 1$}
		\State Select points to be perturbed
		\State $P_{\mathit{O}} \gets P'_{\mathit{O}}$
		\State Insert $P'_O$ into $P'$: $P \gets P'$
		\Else
		\State Insert $P_O$ into $P'$:  $P \gets P'$
		\EndIf
		\EndFor
		\EndFor
		\State $\mathit{iteration} += 1$
		\EndFor
	\end{algorithmic}
\end{algorithm}

\section{Experimental Validation of Six State-of-the-art Object Detection Systems} \label{sec:res}
In the following, we present our experimental results, including the performance of the investigated 3D OD on natural data and the data perturbed by AELiDAR. Thereby, FN\_ASR and FP\_ASR refer to the mean value across all scenes. 
 
\subsection{Hi-ALPS Level 0: Natural Data} \label{sec:resnat}
Table \ref{tab:perfnat} includes the AP values for the classes car, pedestrian and cyclist as well as the mAP values of the investigated pretrained 3D OD models on the natural KITTI validation split, corresponding to Hi-ALPS level 0. Thereby, Part-A² achieves the highest value, while PointPillars achieves the lowest value. Furthermore, we calculated the FN\_ASR and FP\_ASR regarding the predicted objects compared to the ground truth annotations, for all classes as well as all objects without the consideration of the predicted class. Regarding FN\_ASR, the models achieve lower values, i.e., a higher performance, when
regarding all objects without consideration of their class. In contrast, for FP\_ASR, the
values for all objects lie between the class values. In summary, all models perform best for class car and the worst for class pedestrian.

\begin{table*}[t!]
	\centering
	\caption{Performance of state-of-the-art 3D object detection systems on natural KITTI validation split, AP values for different classes and mAP as well as $\\mathit{FN\_ASR}$ and $\mathit{FP\_ASR}$ of the predicted objects compared to the ground truth annotations. }
	\begin{tabular}{l| cccccc}
		\hline
		Class & \multicolumn{6}{c}{Object Detection System} \\ 
		AP&Part-A² & PointPillars & PointRCNN & PV-RCNN & SECOND & VoxSeT \\ \hline 
		&\multicolumn{6}{c}{AP} \\  \hline 
		Car &85.6861 &80.4442 &82.6821 &86.3161 &83.5919 &84.0209 \\
		Pedestrian &60.3971 &51.8625 &55.2595 &55.6937 &51.0836 &54.3185 \\
		Cyclist & 75.8018&67.7353 &75.6413 &75.1663 &70.8295 &75.3128 \\
		mAP &73.9617 &66.6806 &71.1943 &72.3920 &68.5017 &71.2174 \\ \hline
				&\multicolumn{6}{c}{$\mathit{FN\_ASR}$} \\ \hline  
				Car & 0.2697&0.3045 &0.3052 &0.2529 &0.2784 &0.2678 \\
				Pedestrian &0.8145 &0.8530 &0.8137 &0.8178 &0.8332 &0.8258 \\
				Cyclist &0.6757 &0.7537 &0.7096 &0.7041 &0.7182 &0.7343 \\
				Objects &0.1114 &0.1149 &0.1540 &0.1074 &0.1095 &0.1085 \\ \hline
				&\multicolumn{6}{c}{$\mathit{FP\_ASR}$}\\ \hline 
				Car &0.6760 &0.6813 &0.4217 &0.4958 &0.6703 &0.6461 \\
				Pedestrian &0.6731 &0.9560 &0.4657 &0.7339 &0.9152 &0.8889 \\
				Cyclist &0.2874 &0.7756 &0.1879 &0.2537 &0.6436 &0.6212 \\
			Objects &0.6318 &0.7556 &0.3925&0.5232 &0.7059 &0.6898 \\ \hline
	\end{tabular}
	\label{tab:perfnat}
\end{table*}

\subsection{Hi-ALPS levels 1 to 5: Perturbed Data} \label{sec:pertres}
\begin{figure}
	\centering
	\includegraphics[width=0.4 \textwidth]{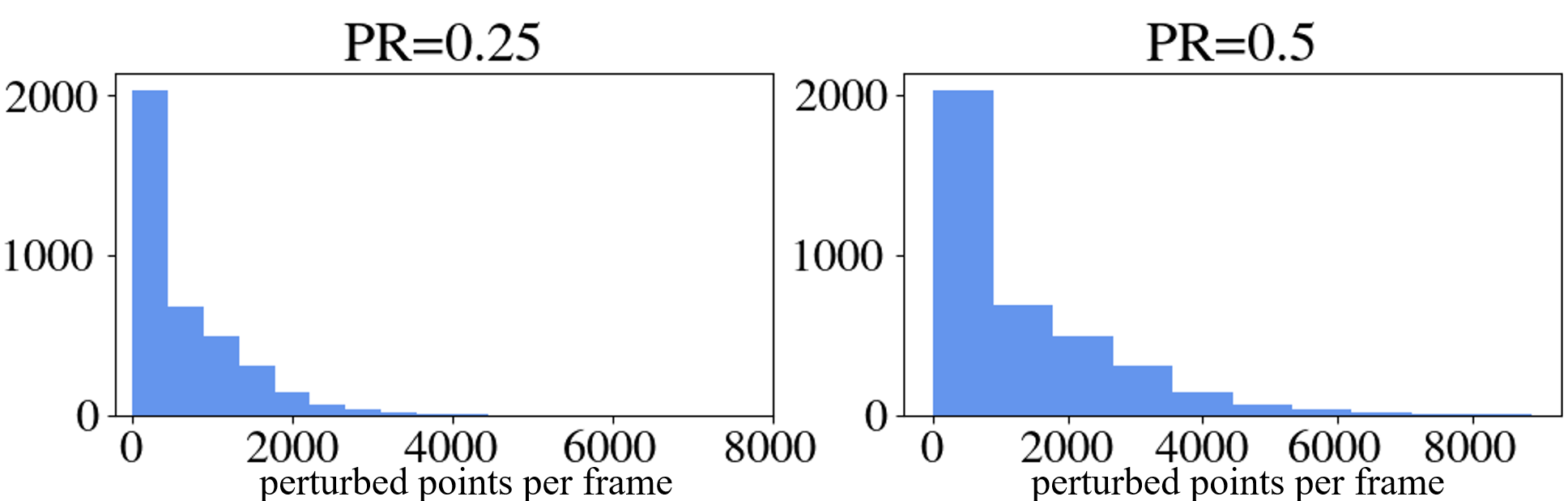}
	\caption{Histograms of the number of perturbed points per LiDAR frame in KITTI dataset for $\mathit{PR}=0.25$ and $\mathit{PR}=0.5$.}
	\label{fig:num_pert}
\end{figure}

As mentioned in Section \ref{sec:pertimpl}, we conduct our experiments in three iterations. The results of these iterations differed insignificantly with a maximal deviation of 3\% between the iteration results.
For our experiments, we chose $\mathit{PR} = \{0.25, 0.5\}$ and $\mathit{SF}= 0.01$.  
Figure \ref{fig:num_pert} shows histograms of the number of perturbed points per LiDAR frame for both PR values. Compared to other publications, \eg, \cite{spoofcao, spoofsun}, the number of perturbed points in our publication is quite high. Nevertheless, Sato \etal \cite{sato} determined a number of 6,000 points as a possible attack capability. For most frames, the number of perturbed points is lower than this value in our experiments. 

\begin{figure} [!t]
	\centering
	\includegraphics[width=0.4\textwidth]{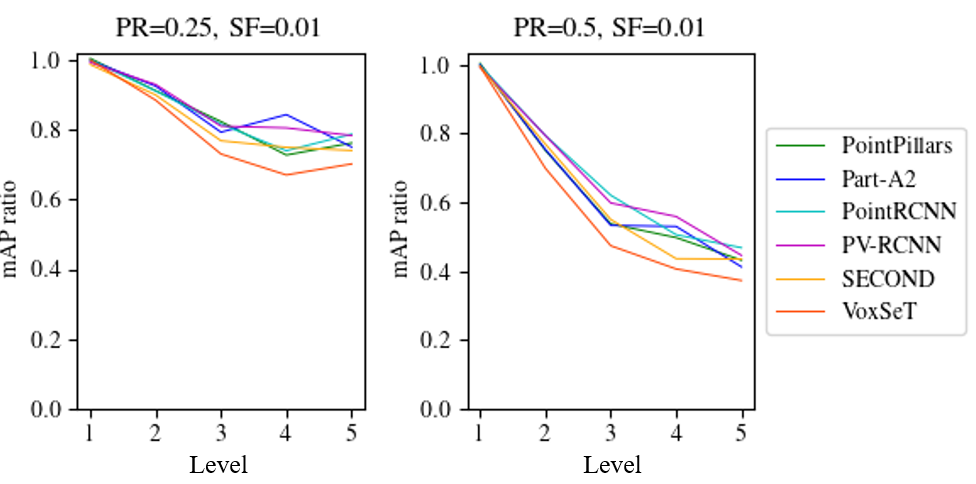}
	\caption{mAP ratio for Hi-ALPS levels 1 to 5. For Hi-ALPS levels 4 and 5, the perturbation type with the highest impact is included.
	}
	\label{fig:map}
\end{figure}

Figure \ref{fig:map} includes the mAP ratio of all investigated OD. For Hi-ALPS levels 4 and 5, the perturbation method with the highest influence on the OD is illustrated. For Hi-ALPS level 4, this is adding points and for Hi-ALPS level 5 dropping points. However, the only exception is PointPillars, which is more vulnerable to point adding at Hi-ALPS level 5, for $\mathit{PR}=0.25$. The figure shows that the higher PR leads to a lower mAP ratio. For $\mathit{PR}=0.25$, the performance decreases for all OD until Hi-ALPS level 3 and then increases for Part-A². For PV-RCNN and SECOND, it does not change significantly. For PointRCNN, PointPillars, and VoxSeT, the mAP ratio increases again from Hi-ALPS level 4 to Hi-ALPS level 5. In contrast, for $\mathit{PR}=0.5$, the mAP ratio decreases for all OD with increasing Hi-ALPS level. Thereby, at Hi-ALPS level 2 it decreases to around 75\% of the value on natural data, and at Hi-ALPS level 5, the performance is even lower than half of the value on natural data. 

\begin{figure}[!t]
	\centering
	\begin{subfigure}{0.235\textwidth}
		\includegraphics[width=\linewidth]{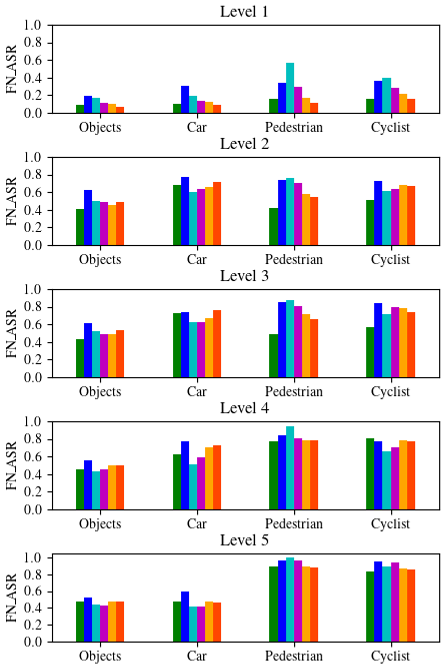}
		\caption{$\mathit{FN\_ASR}$} \label{fig:fn}
	\end{subfigure}
	\begin{subfigure}{0.235\textwidth}
		\includegraphics[width=\linewidth]{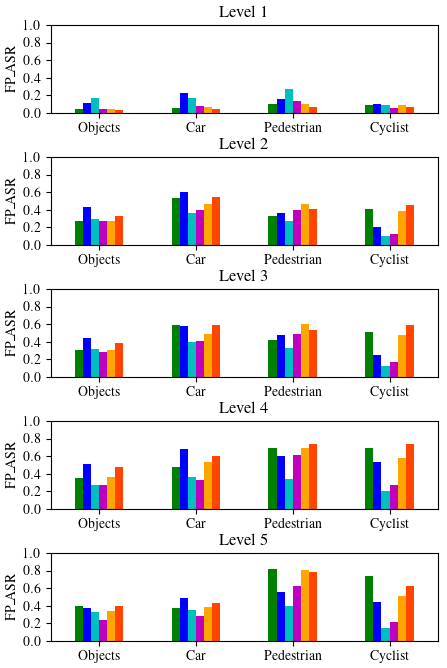}
		\caption{$\mathit{FP\_ASR}$} \label{fig:fp}
	\end{subfigure}
	\begin{subfigure}{0.25\textwidth}
		\includegraphics[width=\linewidth]{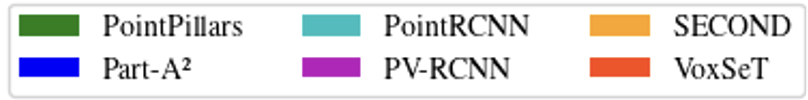}
		\label{fig:l}
	\end{subfigure}
	\caption{$\mathit{FN\_ASR}$ and $\mathit{FP\_ASR}$ for different classes at different Hi-ALPS levels. For Hi-ALPS levels 4 and 5, the perturbation type with the highest impact is included.}
	\label{fig:fnfp}
\end{figure}

Figure \ref{fig:fnfp} shows the results of the investigated OD for the different classes, regarding FN\_ASR and
FP\_ASR. Thereby, \textit{Object} refers to the values of FN\_ASR and FP\_ASR for all objects without consideration of the predicted class. The Figure includes the result of the perturbation method with the highest impact for each class. Table \ref{tab:type} shows the corresponding perturbation method. Green signifies that adding has a higher impact, \ie, the FN\_ASR or FP\_ASR is higher than for dropping, while blue indicates that dropping rather influences the OD. Figure  \ref{fig:fnfp} shows the performance of all OD decreasing from Hi-ALPS level 1 to 3. Thereby, the FN\_ASR and FP\_ASR of the classes pedestrian and cyclist increase stronger than for car and object. PointRCNN tends to achieve rather low values, except for FN\_ASR for class pedestrian, while Part-A² tends to achieve the highest values. 

\begin{table*}[t!]
	\centering
	\caption[Most influential perturbation type for Hi-ALPS level and 5.]{Most influential perturbation type regarding FN\_ASR and FP\_ASR for Hi-ALPS levels 4 and 5, for classes object (O), car (Ca), cyclist (Cy) and pedestrian (P). Green indicates that adding points has more impact, \ie, leads to lower values, while blue means dropping.}
	\begin{tabular}{p{1.7cm}
			|>{\tiny}p{0.05cm} >{\tiny}p{0.05cm} >{\tiny}p{0.05cm} >{\tiny}p{0.05cm}|
			>{\tiny}p{0.05cm}
			>{\tiny}p{0.05cm} >{\tiny}p{0.05cm} >{\tiny}p{0.05cm}|
			>{\tiny}p{0.05cm} >{\tiny}p{0.05cm} >{\tiny}p{0.05cm} >{\tiny}p{0.05cm}|
			>{\tiny}p{0.05cm} >{\tiny}p{0.05cm} >{\tiny}p{0.05cm} >{\tiny}p{0.05cm}|
			>{\tiny}p{0.05cm} >{\tiny}p{0.05cm} >{\tiny}p{0.05cm}
			>{\tiny}p{0.05cm}|
			>{\tiny}p{0.05cm} >{\tiny}p{0.05cm} >{\tiny}p{0.05cm} >{\tiny}p{0.05cm}|
			>{\tiny}p{0.05cm} >{\tiny}p{0.05cm} >{\tiny}p{0.05cm} >{\tiny}p{0.05cm}|
			>{\tiny}p{0.05cm} >{\tiny}p{0.05cm} >{\tiny}p{0.05cm}
			>{\tiny}p{0.05cm}}
	
	\hline
	 	&\multicolumn{16}{c|}{Hi-ALPS level 4} &\multicolumn{16}{c}{Hi-ALPS level 5} \\
	 	
	 	& \multicolumn{8}{c|}{$\mathit{PR=0.25}$}
	 	&\multicolumn{8}{c|}{$\mathit{PR=0.5}$}
	 	& \multicolumn{8}{c|}{$\mathit{PR=0.25}$}
	 	&\multicolumn{8}{c}{$\mathit{PR=0.5}$} \\ 
	 	
		   OD& \multicolumn{4}{c|}{FN\_ASR} & \multicolumn{4}{c|}{FP\_ASR} & \multicolumn{4}{c|}{FN\_ASR} & \multicolumn{4}{c|}{FP\_ASR}& \multicolumn{4}{c|}{FN\_ASR} & \multicolumn{4}{c|}{FP\_ASR}& \multicolumn{4}{c|}{FN\_ASR} & \multicolumn{4}{c}{FP\_ASR}\\

		&O & Ca & P & Cy &
		O & Ca & P & Cy &
		O & Ca & P & Cy &
		O & Ca & P & Cy &
		O & Ca & P & Cy &
		O & Ca & P & Cy &
		O & Ca & P & Cy &
		O & Ca & P & Cy  \\ 
		
		\hline 
	
Part-A²
		 &{\cellcolor{green!25}}
		&{\cellcolor{green!25}}&{\cellcolor{blue!25}}&{\cellcolor{blue!25}}
		
		 &{\cellcolor{green!25}}
		&{\cellcolor{green!25}}&{\cellcolor{green!25}}&{\cellcolor{green!25}}
		
		 &{\cellcolor{green!25}}
		&{\cellcolor{green!25}}&{\cellcolor{blue!25}}&{\cellcolor{blue!25}}
		
		 &{\cellcolor{green!25}}
		&{\cellcolor{green!25}}&{\cellcolor{green!25}}&{\cellcolor{green!25}}
		
	 &{\cellcolor{blue!25}}
	&{\cellcolor{green!25}}&{\cellcolor{blue!25}}&{\cellcolor{blue!25}}
		
		 &{\cellcolor{green!25}}
		&{\cellcolor{green!25}}&{\cellcolor{green!25}}&{\cellcolor{green!25}}
		
		 &{\cellcolor{blue!25}}
		&{\cellcolor{blue!25}}&{\cellcolor{blue!25}}&{\cellcolor{blue!25}}
		
		 &{\cellcolor{green!25}}
		&{\cellcolor{green!25}}&{\cellcolor{green!25}}&{\cellcolor{green!25}}\\
		
PointPillars
		&{\cellcolor{green!25}}
		&{\cellcolor{green!25}}&{\cellcolor{green!25}}&{\cellcolor{blue!25}}
		
		&{\cellcolor{green!25}}
		&{\cellcolor{green!25}}&{\cellcolor{green!25}}&{\cellcolor{green!25}}
		
		&{\cellcolor{blue!25}}
		&{\cellcolor{green!25}}&{\cellcolor{blue!25}}&{\cellcolor{blue!25}}
		
		&{\cellcolor{green!25}}
		&{\cellcolor{green!25}}&{\cellcolor{green!25}}&{\cellcolor{green!25}}
		
		&{\cellcolor{green!25}}
		&{\cellcolor{green!25}}&{\cellcolor{green!25}}&{\cellcolor{green!25}}
		
		&{\cellcolor{green!25}}
		&{\cellcolor{green!25}}&{\cellcolor{green!25}}&{\cellcolor{green!25}}
		
		&{\cellcolor{blue!25}}
		&{\cellcolor{blue!25}}&{\cellcolor{green!25}}&{\cellcolor{blue!25}}
		
		&{\cellcolor{green!25}}
		&{\cellcolor{green!25}}&{\cellcolor{green!25}}&{\cellcolor{green!25}}\\
		
PointRCNN 
	&{\cellcolor{green!25}}
	&{\cellcolor{green!25}}&{\cellcolor{green!25}}&{\cellcolor{green!25}}
	
	&{\cellcolor{green!25}}
	&{\cellcolor{green!25}}&{\cellcolor{green!25}}&{\cellcolor{green!25}}
	
	&{\cellcolor{green!25}}
	&{\cellcolor{green!25}}&{\cellcolor{green!25}}&{\cellcolor{green!25}}
	
	&{\cellcolor{green!25}}
	&{\cellcolor{green!25}}&{\cellcolor{green!25}}&{\cellcolor{green!25}}
	
	&{\cellcolor{blue!25}}
	&{\cellcolor{green!25}}&{\cellcolor{blue!25}}&{\cellcolor{blue!25}}
	
	&{\cellcolor{blue!25}}
	&{\cellcolor{blue!25}}&{\cellcolor{blue!25}}&{\cellcolor{green!25}}
	
	&{\cellcolor{blue!25}}
	&{\cellcolor{blue!25}}&{\cellcolor{blue!25}}&{\cellcolor{blue!25}}
	
	&{\cellcolor{blue!25}}
	&{\cellcolor{blue!25}}&{\cellcolor{blue!25}}&{\cellcolor{green!25}}\\
		
PV-RCNN		
		&{\cellcolor{green!25}}
		&{\cellcolor{green!25}}&{\cellcolor{green!25}}&{\cellcolor{blue!25}}
		
		&{\cellcolor{green!25}}
		&{\cellcolor{green!25}}&{\cellcolor{green!25}}&{\cellcolor{green!25}}
		
		&{\cellcolor{green!25}}
		&{\cellcolor{green!25}}&{\cellcolor{blue!25}}&{\cellcolor{blue!25}}
		
		&{\cellcolor{green!25}}
		&{\cellcolor{green!25}}&{\cellcolor{green!25}}&{\cellcolor{green!25}}
		
		&{\cellcolor{blue!25}}
		&{\cellcolor{green!25}}&{\cellcolor{green!25}}&{\cellcolor{blue!25}}
		
		&{\cellcolor{green!25}}
		&{\cellcolor{green!25}}&{\cellcolor{green!25}}&{\cellcolor{green!25}}
		
		&{\cellcolor{blue!25}}
		&{\cellcolor{blue!25}}&{\cellcolor{blue!25}}&{\cellcolor{blue!25}}
		
		&{\cellcolor{blue!25}}
		&{\cellcolor{blue!25}}&{\cellcolor{green!25}}&{\cellcolor{green!25}}\\
		
SECOND	 	
		&{\cellcolor{green!25}}
		&{\cellcolor{green!25}}&{\cellcolor{green!25}}&{\cellcolor{green!25}}
		
		&{\cellcolor{green!25}}
		&{\cellcolor{green!25}}&{\cellcolor{green!25}}&{\cellcolor{green!25}}
		
		&{\cellcolor{green!25}}
		&{\cellcolor{green!25}}&{\cellcolor{blue!25}}&{\cellcolor{blue!25}}
		
		&{\cellcolor{green!25}}
		&{\cellcolor{green!25}}&{\cellcolor{green!25}}&{\cellcolor{green!25}}
		
		&{\cellcolor{green!25}}
		&{\cellcolor{green!25}}&{\cellcolor{green!25}}&{\cellcolor{green!25}}
		
		&{\cellcolor{green!25}}
		&{\cellcolor{green!25}}&{\cellcolor{green!25}}&{\cellcolor{green!25}}
		
		&{\cellcolor{blue!25}}
		&{\cellcolor{green!25}}&{\cellcolor{blue!25}}&{\cellcolor{blue!25}}
		
		&{\cellcolor{green!25}}
		&{\cellcolor{green!25}}&{\cellcolor{green!25}}&{\cellcolor{green!25}}\\
		
		VoxSeT		
		&{\cellcolor{green!25}}
		&{\cellcolor{green!25}}&{\cellcolor{green!25}}&{\cellcolor{green!25}}
		
		&{\cellcolor{green!25}}
		&{\cellcolor{green!25}}&{\cellcolor{green!25}}&{\cellcolor{green!25}}
		
		&{\cellcolor{green!25}}
		&{\cellcolor{green!25}}&{\cellcolor{green!25}}&{\cellcolor{blue!25}}
		
		&{\cellcolor{green!25}}
		&{\cellcolor{green!25}}&{\cellcolor{green!25}}&{\cellcolor{green!25}}
		
		&{\cellcolor{green!25}}
		&{\cellcolor{green!25}}&{\cellcolor{green!25}}&{\cellcolor{green!25}}
		
		&{\cellcolor{green!25}}
		&{\cellcolor{green!25}}&{\cellcolor{green!25}}&{\cellcolor{green!25}}
		
		&{\cellcolor{blue!25}}
		&{\cellcolor{blue!25}}&{\cellcolor{green!25}}&{\cellcolor{blue!25}}
		
		&{\cellcolor{green!25}}
		&{\cellcolor{green!25}}&{\cellcolor{green!25}}&{\cellcolor{green!25}}\\
		
	 \hline
	\end{tabular}
	\label{tab:type}
\end{table*}

The results described above show that the performance of all OD changes significantly at Hi-ALPS level 3 (Figures \ref{fig:map} and \ref{fig:fnfp}). To investigate the variation of the robustness as well as perturbation metrics, Figure \ref{fig:histos} shows histograms of how often metrics in different value intervals appear. Figure \ref{fig:histofnfp} includes the FN\_ASR and FP\_ASR values across all iterations and LiDAR frames for Hi-ALPS level 3 for all objects. All OD, except Part-A², achieve rather higher FN\_ASR values than the mean value. For FP\_ASR, all OD achieve rather lower values, except Part-A² and VoxSeT.

\subsection{Hi-ALPS Level 3: Consideration of the Environment} \label{sec:env}
\begin{figure} [!t]
	\centering
	\includegraphics[width=0.5\textwidth]{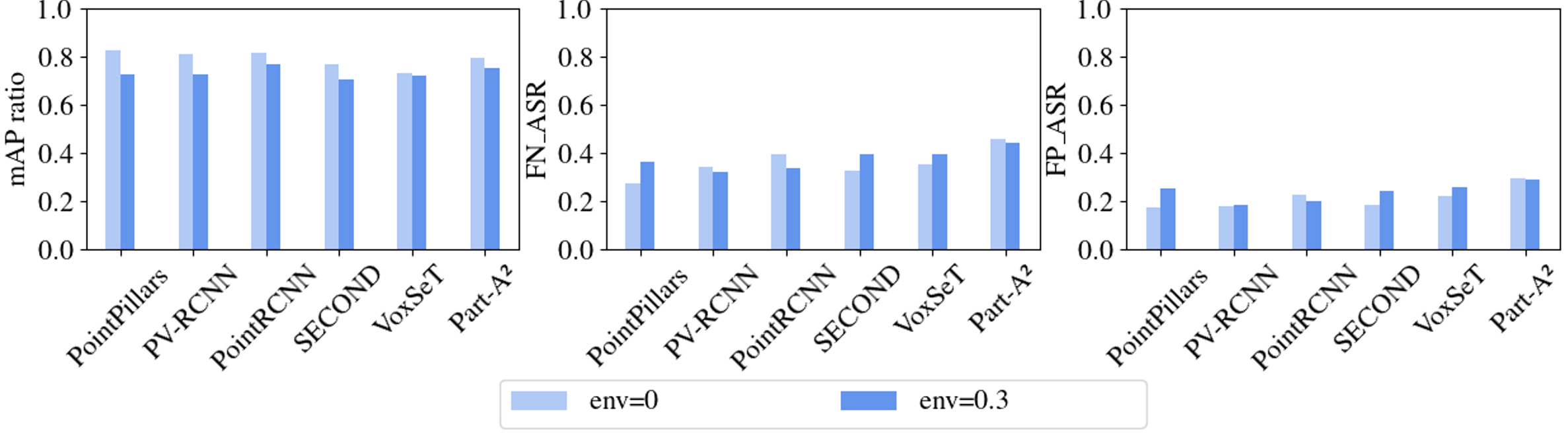}
	\caption{Evaluation with additional environment, mAP ratio, $ \mathit {FN\_ASR}$ and $\mathit {FP\_ASR}$ for all objects without consideration of the predicted class for Hi-ALPS level 3, $\mathit{PR}=0.25$.}
	\label{fig:env}
\end{figure}

To investigate the influence of the environment around each object, we add an additional area around the BB (Section \ref{sec:pertimpl}) during the generation of Hi-ALPS level 3. We select Hi-ALPS level 3 for this experiment, because it influenced the performance of all OD significantly (Figure \ref{fig:map}).  Figure \ref{fig:env} shows the results with the environment. Since adding an environment may increase the influence of the perturbations, we set the PR to $\mathit{PR}=0.25$.  Thereby, the environment decreases the mAP ratio for all investigated OD. The FN\_ASR increases for PointPillars, SECOND and VoxSeT. The FP\_ASR increases for all OD, except for PointRCNN and Part-A².

\subsection{Perturbation Measurement}
\begin{figure}[t!]
	\centering
	\begin{subfigure}[b]{0.24\textwidth}
		\includegraphics[width=\linewidth]{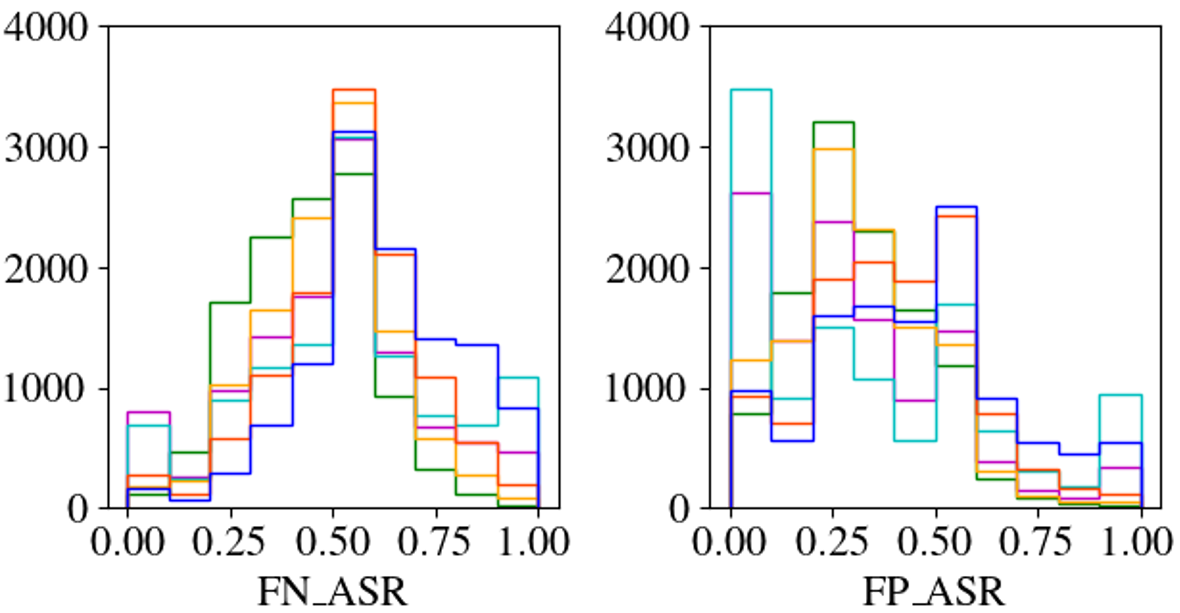}
		\caption{FN\_ASR and FP\_ASR at Hi-ALPS level 3.}
		\label{fig:histofnfp}
	\end{subfigure}
	\begin{subfigure}[b]{0.23\textwidth}
		\includegraphics[width=\linewidth]{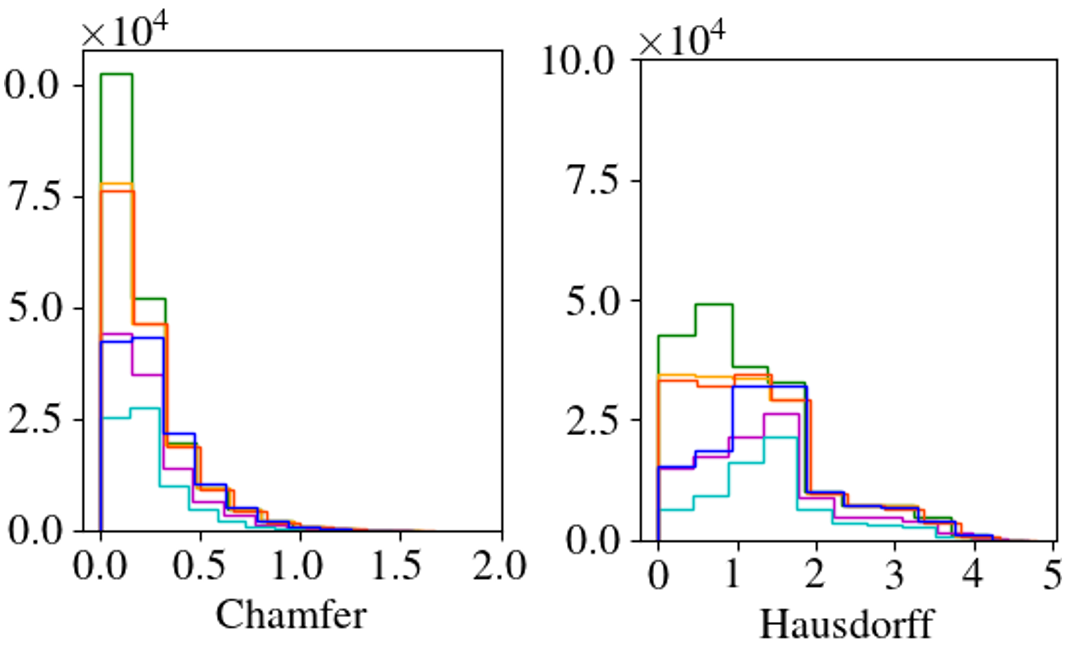}
		\caption{Chamfer and Hausdorff distance for Hi-ALPS level 4 'add'.}
		\label{fig:histopert}
	\end{subfigure}
	\includegraphics[width=0.4\linewidth]{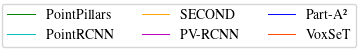}
	\caption{Histograms of robustness and perturbation metrics across all iterations and LiDAR frames for $\mathit{PR}=0.5$.}
	\label{fig:histos}
\end{figure}

\begin{figure}[t!]
	\centering
	\begin{subfigure}{0.241\textwidth}
		\includegraphics[width=\linewidth]{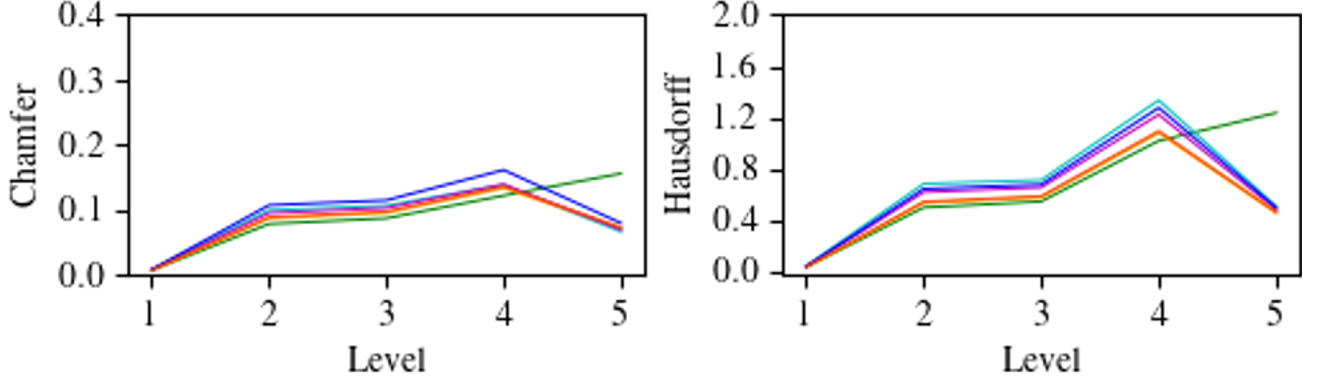}
		\caption{$\mathit{PR} = 0.25$} \label{fig:pr025}
	\end{subfigure}
	\begin{subfigure}{0.241\textwidth}
		\includegraphics[width=\linewidth]{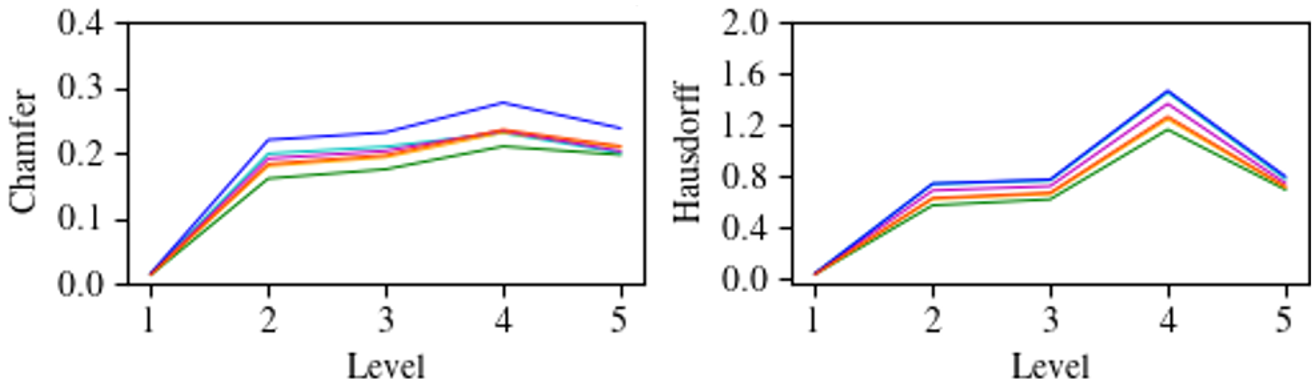}
		\caption{$\mathit{PR} = 0.5$} \label{fig:pr05}
	\end{subfigure}
	
	\includegraphics[width=0.4\linewidth]{figures/legend_line}
	\caption{Mean values of Chamfer and Hausdorff distance for all perturbed point cloud objects. For Hi-ALPS levels 4 and 5, the most influential perturbation method regarding mAP ratio is included, this is 'add' for Hi-ALPS level 4 and 'drop' for Hi-ALPS level 5, except for PointPillars, $\mathit{PR}=0.25$, where it is 'add'. }
	\label{fig:pertlevel}
\end{figure}

Figure \ref{fig:pertlevel} includes the mean values of Chamfer and Hausdorff distance for all perturbed point objects of interest included in the dataset. For Hi-ALPS levels 4 and 5, the values of the most influential perturbation types are included, regarding the mAP ratio, because it continuously decreases with increasing Hi-ALPS level (Figure \ref{fig:map}). Thereby, the higher the PR, the higher are Chamfer and Hausdorff distance.  
The values increase until Hi-ALPS level 4 and decrease for  \mbox{Hi-ALPS level 5}, except for PointPillars, where the values further increase at Hi-ALPS level 5 for $\mathit{PR}=0.25$. 
In summary, for Part-A² and PointRCNN, AELiDAR achieves the highest Chamfer and Hausdorff distance. In contrast, the values are the lowest for PointPillars, except for \mbox{Hi-ALPS level 5}, $\mathit{PR}=0.5$.
Figure \ref{fig:histopert} includes histograms of the Chamfer and Hausdorff distance values for all perturbed point objects in all iterations for Hi-ALPS level 4 when adding points for $\mathit{PR}=0.5$, because the perturbations of this Hi-ALPS level include the most degrees of freedom. For all OD, the values of the perturbation metrics are rather lower than the mean.

\section{Discussion}
In this section, we discuss our experimental results. First, we conduct a robustness analysis and afterwards, we discuss the perceptibility of the computed perturbations. Finally, we discuss the real-world plausibility of our perturbations.

\subsection{Object Detection System Robustness Analysis}
In the following, the results of the OD on the perturbations included in Hi-ALPS are discussed. 
We refer to the results for $\mathit{PR} = 0.5$, because this configuration stronger influences the OD, \ie, it leads to lower mAP ratio values and higher FN\_ASR and FP\_ASR values.

\paragraph{Hi-ALPS Level 0: Natural Data}
The different OD achieve a different performance on natural data regarding different metrics, \eg,  Part-A² achieves the highest mAP value and, therefore, the best performance regarding mAP, while PV-RCNN performs the best regarding FN\_ASR, \ie, it achieves the lowest value. This indicates the importance of considering multiple
metrics in the discussion of the performance of the OD. Thereby, the AP provides an overview of the performance, while FN\_ASR and FP\_ASR refer to overlooked objects and false positive detections.
Note that the FP\_ASR values on natural data are rather high. This is due to the KITTI annotations, which only include labels for objects visible in the image plane, where non-visible objects are not considered in the evaluation \cite{kittibench}. The high FP\_ASR values in context with the  AP values indicate that most of the false positives are not considered. The prediction score thresholds of the OD to consider an object as detected are set to a value of 0.1. Therefore, the OD are very sensitive to positive predictions. However, a high FP\_ASR is not as unfavorable as high FN\_ASR values, since it suggests a higher trustworthiness.
 The results show that the OD perform differently regarding different classes. All investigated OD perform the best for the class car, while at the same time, the classes pedestrian and cyclist are rather exposed since these do not have a protective bodywork. This indicates the need to further improve the performance of the OD on natural data. 

\paragraph{Hi-ALPS Levels 1 to 5: Robustness Against the Analyzed Perturbations}
The mAP ratio decreases for all OD with increasing Hi-ALPS level, regardless of their category \mbox{(Table \ref{tab:odcat})}. Thereby, Hi-ALPS level 1 did not significantly influence any of the investigated OD. 
For the following Hi-ALPS levels, point-based PointRCNN and point-voxel-based PV-RCNN, both two-stage architectures, tend to achieve  the highest mAP ratio values. 
Regarding the class-specific robustness, based on Figure \ref{fig:fnfp}, according to FN\_ASR and FP\_ASR, the class pedestrian is the most vulnerable. This is because pedestrians include the smallest number of points, because of their size. Therefore, they are in particular vulnerable to perturbations. 
However, the class car is vulnerable to the perturbations as well, even though it tends to include the largest amount of points. This may be because car is the most frequent class in the KITTI dataset \cite{kitti}. Therefore, the OD may be overfitted to this class, making them especially sensitive to perturbations in objects belonging to the class car.
Regarding the results on natural data, due to the high number of false positive detections, the high FN\_ASR value might indicate an increasing performance, since the overlooked objects could theoretically address false positive detections on natural data that were not detected on perturbed data anymore. However, the decreasing mAP values indicate a decreasing performance making this assumption unlikely.
According to all results, there is no significant difference notable between different categories of OD, with respect to representation and architecture (Section \ref{sec:3dod}).

Regarding the influence of different perturbation types, for Hi-ALPS level 4, which perturbs the point number at random locations of the object, in terms of mAP ratio and FP\_ASR, all OD are rather influenced by adding points than dropping points. This is because the high number of included points in the  point cloud objects results in a high information density, where not all points are relevant to define the object. Therefore, the randomly dropped points do not necessarily correspond to the points which influence the OD. 
For Hi-ALPS level 5, regarding AP ratio and FN\_ASR, the OD are mainly influenced by point dropping, since the outer points are removed, \ie, the contour of the objects is perturbed. For FN\_ASR as well, point dropping mostly leads to the highest values. 
Regarding FP\_ASR, point adding, \ie, increasing the number of included points and the size of the object, influences the OD the most, because more points may indicate to the OD that there are larger objects. Consequently, the predicted BB on the perturbed data may be larger and, therefore, the IoU threshold with the natural prediction may not be fulfilled. In addition, when adding points into the empty spaces inside the bounding box, the added points may be accidentally detected as additional objects.

\paragraph{Hi-ALPS level 3: Influence of the Environment}
The results show that the consideration of the environment during the generation of the perturbations influences the performance of 3D OD (see Section \ref{sec:env}). This is because the environment may include context information about the object, such as its position in the scene or occlusions. Furthermore, it increases the value of the perturbation metrics, because due to the larger BB the maximum shift distance increases. 

\subsection{Perceptibility of the Perturbations}
\begin{figure*}[!t]
	\centering
	\begin{subfigure}{0.32\textwidth}
		\includegraphics[width=\linewidth]{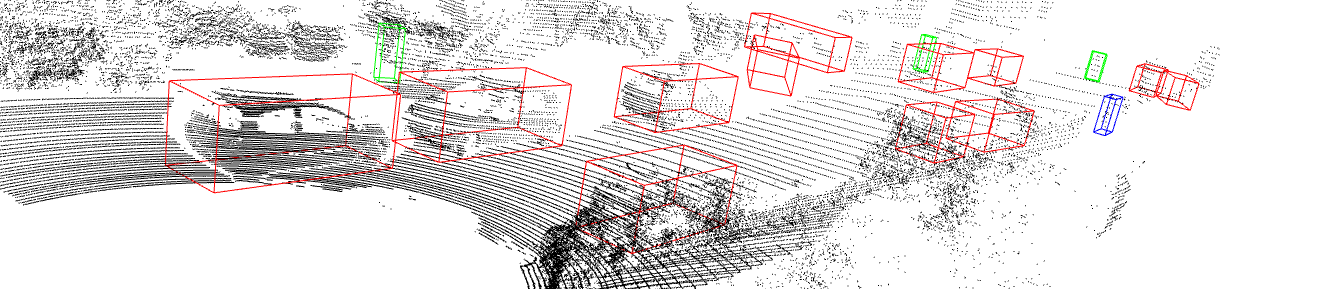}
		\caption{Natural scene} \label{fig:1apart}
	\end{subfigure}
	\begin{subfigure}{0.32\textwidth}
		\includegraphics[width=\linewidth]{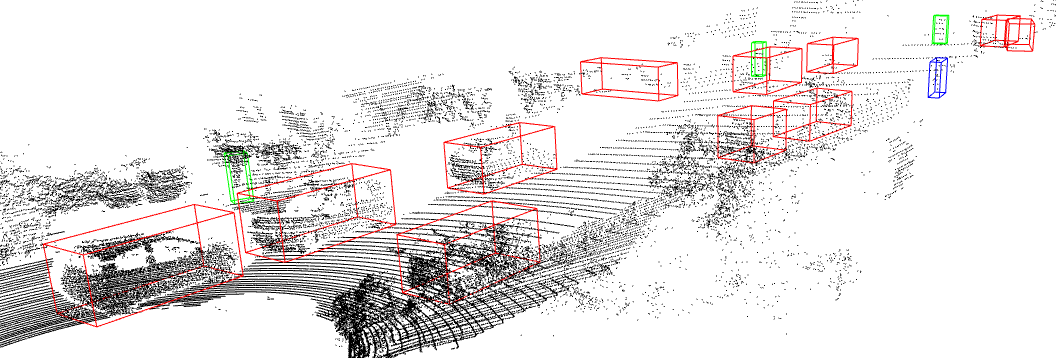}
		\caption{Hi-ALPS level 1} \label{fig:1bpart}
	\end{subfigure}
	\begin{subfigure}{0.32\textwidth}
		\includegraphics[width=\linewidth]{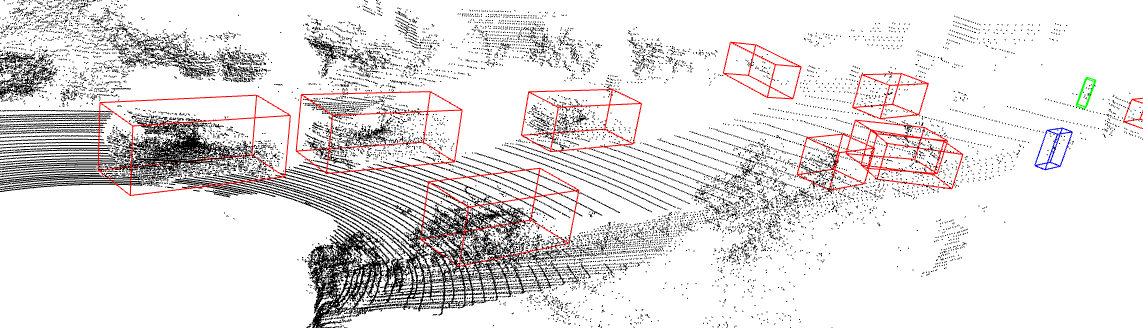}
		\caption{Hi-ALPS level 2} \label{fig:1cpart}
	\end{subfigure}
	\begin{subfigure}{0.32\textwidth}
		\includegraphics[width=\linewidth]{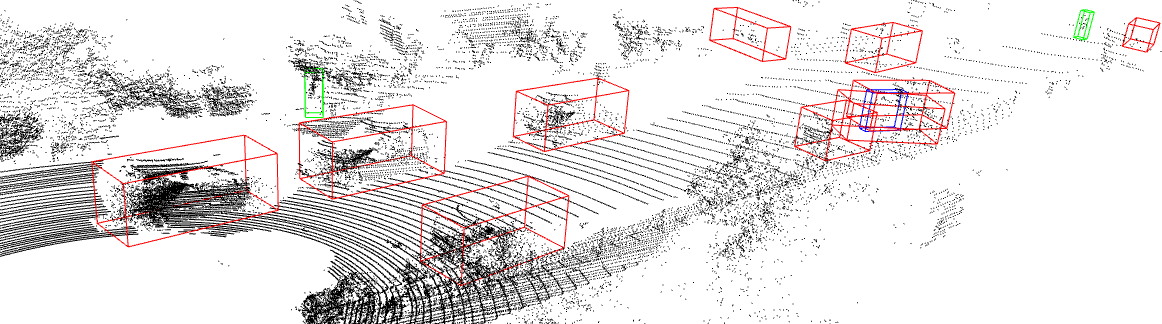}
		\caption{Hi-ALPS level 3} \label{fig:1dpart}
	\end{subfigure} 
	\begin{subfigure}{0.32\textwidth}
		\includegraphics[width=\linewidth]{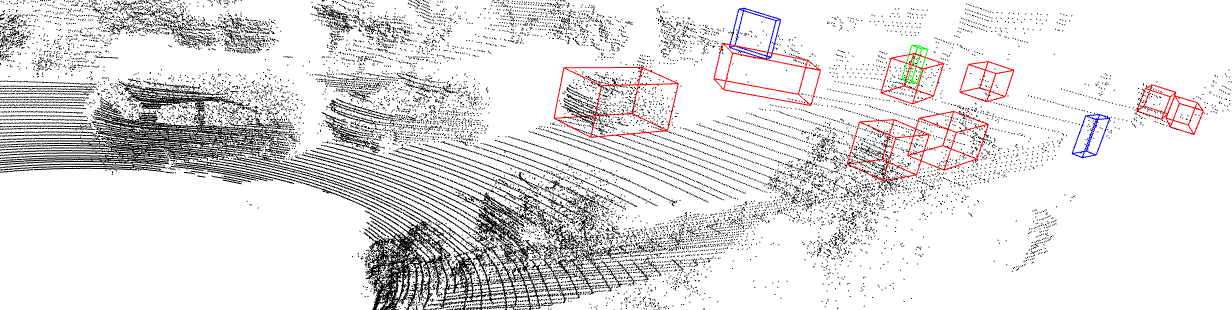}
		\caption{Hi-ALPS level 4 'add'} \label{fig:1epart}
	\end{subfigure}
	\begin{subfigure}{0.32\textwidth}
		\includegraphics[width=\linewidth]{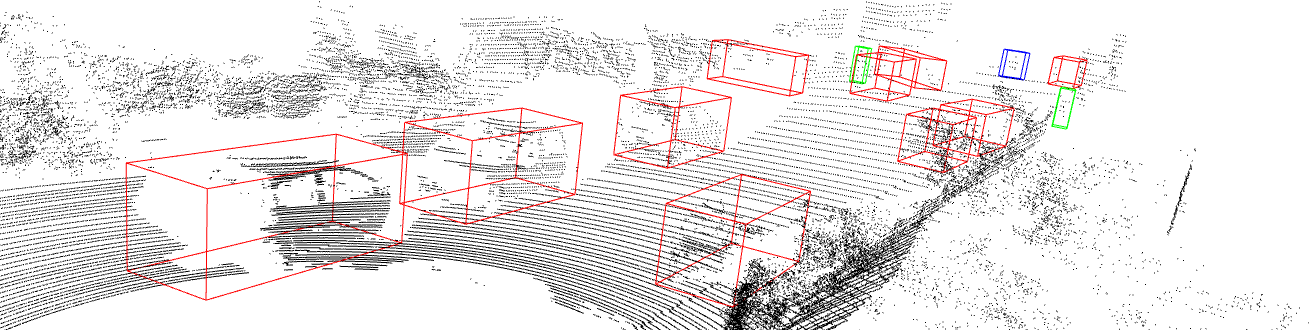}
		\caption{Hi-ALPS level 5 'drop'} \label{fig:1fpart}
	\end{subfigure}
	
	\caption{Hi-ALPS level 1 to 5, including the predictions of Part-A², $\mathit{PR}=0.5$. Red boxes refer to class car, blue to cyclist and green to pedestrian.}
	\label{fig:parta2}
\end{figure*}

The following discusses the perceptibility of the generated perturbations, because their imperceptibility is a basic requirement for them to be considered as adversarial examples (Paragraph \ref{sec:ae}). 
 The values of Chamfer and Hausdorff distance are similar for all investigated OD but differ, because the BB predicted by different OD are not the same size. For Hi-ALPS levels 2 and 3, the shift distance is not limited by the SF but by the center of the BB instead. Larger BB lead to larger allowed shift distances (Section \ref{sec:pertimpl}). The perturbation metrics further increase for Hi-ALPS level 4, since the distance of the added points to the natural points is solely limited by the BB. For Hi-ALPS level 5, both metrics decrease again, because instead of adding points, AELiDAR drops natural points. Therefore, the distances between existing points change, while adding new points rather leads to outliers and greater distances. For the smaller $\mathit{PR}=0.25$, the perturbation metrics of PointPillars increase at Hi-ALPS level 5. This is, because with $\mathit{PR}=0.25$, PointPillars is more sensitive to adding points instead of dropping points. Furthermore, according to Figure \ref{fig:histopert}, the values of the perturbation metrics are rather lower than the mean values, \ie, the perturbations are rather imperceptible than the values the perturbation metrics suggest. The higher Hausdorff than Chamfer distance values suggest this as well.  
Figure \ref{fig:parta2} visualizes a point cloud scene 
with the predictions of Part-A², $\mathit{PR}=0.5$. Thereby, in all Hi-ALPS levels, the objects are still recognizable to a human observer. At Hi-ALPS level 4 (Figure \ref{fig:1epart}), Part-A² does not detect the car in the front, while at Hi-ALPS level 5 (Figure \ref{fig:1fpart}), it detects this car, even though, half of it is missing. 

\subsection{Summary of the Experimental Validation}
Although our perturbations are not generated using sophisticated adversarial example algorithms---\ie, they do not constitute strong attacks and are therefore less effective than state-of-the-art methods (\eg, \cite{scenesurvey, spoofsun, sato})---the robustness of the OD is insufficient for any of the Hi-ALPS levels, except for Hi-ALPS level 1. 
 There is no clear correlation between the category an OD belongs to and its robustness against the perturbations. Even though the perturbation metrics are increasing and higher than the values presented in other publications, \eg, \cite{scenesurvey, minimal}, a human observer can still correctly identify the objects and, therefore, the investigated OD should as well.  Furthermore, since with increasing Hi-ALPS level the robustness of the OD decreases, Hi-ALPS is internally coherent. 

\subsection{Real-world Plausibility}
In this section, we discuss the real-world plausibility of the perturbations included in the Hi-ALPS levels (Section \ref{sec:pertlevel}), addressing natural events (safety) as well as attacks conducted by a malicious attacker (security). 

\subsubsection{Shifting}
Regarding safety, shifted points may occur in the real world caused by various reasons, \eg, limitations of the LiDAR sensor system, \eg, sensor vibrations \cite{scenesurvey, robo3d}. Another reason may be the movement of objects \cite{scenesurvey, robo3d, dong}. Since our attacks address single point cloud objects, which correspond to humans, the movement of these may lead to shifted points. Consequently, it cannot be ruled out that our perturbations generated in Hi-ALPS levels 1 to 3 may occur in the real world.

Regarding security, a malicious attacker may implement shifted points by moving the LiDAR sensor, \eg, hitting it, or with moving adversarial objects. However, shifting the most influential points so that these change the prediction is probably rather complex to implement. 

\subsubsection{Adding}
Regarding safety, examples of natural events causing additional points are fog, which causes backscattered points, or laser beams from other LiDAR-equipped autonomous vehicles. 

Regarding security, in particular, random points, such as  in Hi-ALPS level 4, can easily be spoofed (Paragraph \ref{sec:ae}). Thereby, according to our experimental results (Section \ref{sec:pertres}), for a successful attack, the attacker does not even have to determine suitable locations for the additional points. Instead, already points randomly spoofed nearby an object, corresponding to Hi-ALPS level 4, decrease the performance of the OD significantly (Figure \ref{fig:1epart}). Furthermore, Sato \etal \cite{sato} were able to spoof 6,000 points. As shown in Figure \ref{fig:num_pert}, the number of added points in our experiments is lower than this value for almost all LiDAR frames. 
However, addressing all objects in a frame as in our experiments remains challenging, since Sato \etal determined an angle of 80° to be the maximum spoofing capability. Nevertheless, even spoofing points to perturb one single false object within the frame can have severe real-world consequences and this approach is significantly easier to implement in practice.

\subsubsection{Dropping}
Regarding safety, point dropping is likely to happen through occlusions in the real world. These can be caused by different weather conditions, \eg, strong snowfall, where single points of an object may be missing in the point cloud data. This mainly addresses random points, such as Hi-ALPS level 4. However, weather events address the whole scene instead of single point cloud objects. In contrast, in Hi-ALPS level 5, points in particular parts of the point cloud object are dropped. This is rather likely to happen in the real world, since the occlusion can happen due to just one object disturbing the LiDAR, \eg, an unfortunately passing leaf covering half of a car could lead to the scenario illustrated in Figure \ref{fig:1fpart}.

Regarding security, a malicious attacker can cause dropped points as well, such as by manipulating the LiDAR sensor, \eg, by covering parts of its surface with paint,  or by using surfaces with unfavorable reflectivity. 


\section{Potential Countermeasures to Increase the Robustness Against the Analyzed Perturbations}

In this section, we discuss potential countermeasures to increase the robustness of the OD against the investigated perturbations. 
First, we discuss the applicability of state-of-the-art countermeasures from the literature and then derive further countermeasures based on our experimental results.

\subsection{Discussion of State-of-the-art Countermeasures}
The following discusses the applicability of countermeasures presented in Paragraph \ref{sec:ae} for improving the robustness of the OD against the investigated perturbations.

\paragraph{Adversarial Training}
Adversarial training involves incorporating correctly labeled adversarial data into the training process, thereby teaching the OD to correctly interpret perturbed data. Therefore, it is likely an effective countermeasure against all Hi-ALPS levels. However, since the perturbations in Hi-ALPS are random, it is difficult to cover all possible variations of these within the adversarial training process. Furthermore, adversarial training increases the time as well as computational resources required for training.

\paragraph{Input Transformation}
Simple random sampling, statistical outlier removal, salient point removal,  and DUP-Net, are applied to the input data of the OD and remove points to denoise the perturbed point cloud data. Therefore, these methods are in particular likely to increase the robustness against Hi-ALPS levels, which generate additional points, \ie, Hi-ALPS levels 4 and 5. Thereby, the results of Hi-ALPS level 5 in Figure \ref{fig:1fpart} indicate that even the removal of a high number of natural points does not necessarily influence the OD.  Furthermore, these methods might be effective against Hi-ALPS level 1 as well, since in particular statistical outlier removal, salient point removal, and DUP-Net based on statistical outlier removal are likely to remove the perturbed points. In contrast, for Hi-ALPS levels 2 and 3, these countermeasures are probably less effective, since the contour of the point cloud object is already damaged by the perturbations and these countermeasures could potentially exacerbate the perturbations. 

Nevertheless, input transformations, as an additional preprocessing step that modifies the input data, carry the risk of behaving like perturbations, potentially decreasing the performance of the OD. This applies in particular for simple random sampling, as this method removes points randomly, which may not correspond to actual perturbed points. Statistical outlier removal can determine natural points as outliers and mistakenly remove them. In the case of salient point removal, if no significantly influential perturbed points exist, this method may act like a white-box adversarial attack algorithm. Moreover, it requires access to the OD's gradient. Finally, the upsampling process in DUP-Net may need to consider the LiDAR sensor's scanning pattern, specifically the horizontal lines visible in the point cloud data (e.g., Figure \ref{fig:parta2}).

\paragraph{Countermeasures Exploiting Physical Properties}
Countermeasures exploiting physical properties, such as Yang \etal \cite{yang} or CARLO and SVF, both developed by Sun \etal \cite{spoofsun}, are probably especially effective against Hi-ALPS level 4 and 5, when adding points. Based on the physical properties CARLO exploits, it is potentially capable of detecting shifted points (Hi-ALPS levels 1 to 3), as well as the countermeasure of Yang \etal. In contrast, SVF may increase the robustness of the OD against Hi-ALPS level 1 but not for Hi-ALPS levels 2 and 3, since these shift points towards the center of the point cloud object. Furthermore, SVF is probably not a suitable countermeasure for dropping points, as included in Hi-ALPS levels 4 and 5. This is because dropped points are not visible in the front view in contrast to shifted or added points. 
Since these countermeasures consider physical laws, in contrast to classical input transformations, they are less likely to act like perturbations. Nevertheless, their implementation is more complex than that of classical input transformations, such as Statistical Outlier Removal.

\subsection{Derivation of Potential Countermeasures}
In the following, we propose potential countermeasures derived from our experimental results (Section \ref{sec:res}) for improving the performance of the investigated OD. These include model-focused approaches as well as input transformation.

\subsubsection{Model-focused}
We propose model-focused countermeasures derived from the experimental results. Their derivation is based on the observation of the behavior of the investigated OD on natural as well as on perturbed data. 

\paragraph{Prediction Score Threshold}
According to the results on natural data (Section \ref{sec:resnat}), the OD tend to predict a lot of false positives. One reason for this might be the prediction score threshold, which is set to a potentially too low value and may result in too sensitive OD.  Therefore, a potential deep model modification could lie in adapting the prediction score threshold to balance it based on FN\_ASR and FP\_ASR. Thereby, different prediction score thresholds might be required for different classes. However, this modification requires an extensive study, including the optimization and evaluation of the prediction score threshold. 

\paragraph{Model Retraining}
All OD perform the best for the class car, which is the most frequent class in the KITTI dataset, and the worst for class pedestrian. Data augmentation may be an effective approach to increase performance by retraining the OD  for the classes pedestrian and cyclist. This can be achieved by extending the dataset with more scenes in particular including pedestrians and cyclists. Alternatively, a more balanced dataset can be achieved by removing scenes that include exclusively cars. The second option is easier to realize, however, it reduces the amount of train data and could lead to a general decrease in the  performance of the OD.

\subsubsection{Input Transformation}
We propose approaches for restoring the natural point cloud data. These suggestions are derived based on the visualization of the perturbed data. 

\paragraph{Thinning out Point Clusters}
In most cases, there is an empty space inside the predicted BB, since the objects do not correspond to cuboids (see Figure \ref{fig:parta2}). In Hi-ALPS levels 4 and 5, AELiDAR adds additional points into these naturally empty spaces. Consequently, a potential countermeasure could lie in thinning out these point clusters by dropping points in the top corners of the BB to restore the object. This countermeasure might be effective as well against Hi-ALPS level 1, since it may shift points into the naturally empty spaces.  Since Hi-ALPS levels 2 and 3 shift points into the point cloud center, thinning it out might potentially improve the robustness as well. Against dropping points, this approach is probably not effective, since it drops further points. However, this approach requires the identification of the BB or of the natural empty spaces described above to restore the data.

\paragraph{Generative Approaches}
The violation of physical laws offers the identification of perturbations. Occlusions result in naturally empty spaces. In contrast, point dropping leads to empty spaces without points belonging to objects occluding these (see Figure \ref{fig:1fpart}). Thereby, the size of these naturally not empty spaces and the scanning pattern may offer a basis for reconstructing the missing area of the point cloud. The reconstruction can potentially be conducted as a preprocessing step of the input data by generative approaches, such as autoencoders or GAN. This may be an effective countermeasure for dropping points and potentially shifting points, since these may result in empty spaces. However, the challenges lie in compliance with the physical law and the need for large amounts of diverse training data.

\section{Conclusion}
In this paper, we analyze the robustness of six state-of-the-art 3D OD and quantify them based on the hierarchical perturbation level system Hi-ALPS. These six 3D OD cover different categories based on representation and architectures for LiDAR data analyses established in autonomous driving and, therefore, constitute a representative spectrum of 3D OD. In a series of comprehensive experiments, we demonstrate that the hierarchical perturbation level system comprises consistent, \ie, increasingly complex perturbation levels that can be used as reference levels for the quantification of robustness. In addition, we analyze the six 3D OD based on Hi-ALPS and found that none of them is robust across all examined levels of perturbation. Already at Hi-ALPS level 2, all six OD cannot fully maintain their performance. The analyzed perturbations reduce the performance of the 
OD, \ie, at Hi-ALPS level 2 to only about 75\%, and at Hi-ALPS level 5 even to less than half compared to the performance on natural data. The performance drops already on the second level shows that the analyzed OD are not robust, as the analyzed perturbations are exclusively based on a heuristic, and, therefore, do the generated perturbations definitely not correspond to a strong attack, against which the OD have no chance. The failure of the state-of-the-art OD already at the second perturbation level indicates that the training of the OD needs to be optimized, \eg, with more diverse data, such as \cite{robo3d, dong, robuli}, before complex countermeasures would need to be applied. However, we discuss the applicability of potential countermeasures from related work and propose further countermeasures based on our experimental validation to increase the robustness against higher perturbation levels in the future.

\IEEEtriggeratref{26}
\bibliographystyle{IEEEtran}
\bibliography{main}


\end{document}